\PassOptionsToPackage{table,dvipsnames}{xcolor}
\documentclass{article} 

\usepackage{iclr2026_conference,times}
\iclrfinalcopy   

\usepackage{amsmath,amsfonts,bm}

\def\eqref#1{equation~\ref{#1}}

\def\1{\bm{1}}

\DeclareMathAlphabet{\mathsfit}{\encodingdefault}{\sfdefault}{m}{sl}
\SetMathAlphabet{\mathsfit}{bold}{\encodingdefault}{\sfdefault}{bx}{n}

\usepackage{hyperref}
\usepackage{url}
\usepackage{graphicx}
\usepackage{booktabs}
\usepackage{multirow}
\usepackage{makecell}
\usepackage{subcaption}
\usepackage{tabularx}
\usepackage{array}
\usepackage{enumitem}
\usepackage{pifont}
\usepackage{arydshln}
\usepackage{float}
\usepackage{graphicx}
\usepackage{subcaption}
\usepackage{tikz}
\usepackage{xcolor} 
\usepackage{colortbl}
\usepackage{tabularx}
\usepackage{ragged2e}

\usepackage{tcolorbox}
\tcbuselibrary{listings,breakable,skins} 
\usepackage[utf8]{inputenc}   
\usepackage[T1]{fontenc}      
\usepackage{inconsolata}      

\newcolumntype{Y}{>{\raggedright\arraybackslash\hspace{0pt}}X}
\renewcommand{\arraystretch}{1.06}
\newtcblisting{promptlist}[1][]{%
  listing only,
  breakable,
  colback=gray!5,
  colframe=black!60,
  boxrule=0.5pt,
  listing options={
    basicstyle=\ttfamily\small,
    breaklines=true,
    columns=fullflexible,
    keepspaces=true,
    showstringspaces=false,
    upquote=true
  },
  #1 
}

\definecolor{HeaderGray}{HTML}{F2F2F2}   
\definecolor{OursBlue}{HTML}{E8F2FF}     

\title{Discerning Minds or Generic Tutors? \\Evaluating Instructional Guidance Capabilities in Socratic LLMs}

\author{Ying Liu,~ Can Li, ~Ting Zhang\thanks{Corresponding Author.}~,~ Mei Wang, ~Qiannan Zhu, ~Jian Li, ~Hua Huang \\
Beijing Normal University
}

\begin{document}

\maketitle

\begin{abstract}
The conversational capabilities of large language models hold significant promise for enabling scalable and interactive tutoring. 
While prior research has primarily examined their ability to generate Socratic questions, it often overlooks a critical aspect: adaptively guiding learners in accordance with their cognitive states.
This study moves beyond question generation to emphasize instructional guidance capability. We ask: Can LLMs emulate expert tutors who dynamically adjust strategies in response to learners' states?
To investigate this, we propose GuideEval, a benchmark grounded in authentic educational dialogues that evaluates pedagogical guidance through a three-phase behavioral framework: (1) Perception, inferring learner states; (2) Orchestration, adapting instructional strategies; and (3) Elicitation, stimulating proper reflections.
Empirical results indicate that existing LLMs often fail to provide effective adaptive scaffolding when learners experience confusion or require redirection. To complement the quantitative evaluation, we conduct a detailed failure case analysis, providing an intuitive understanding of these shortcomings. Furthermore, we introduce a behavior-guided finetuning strategy that leverages behavior-prompted instructional dialogues, substantially enhancing guidance performance.
By shifting the focus from isolated content evaluation to learner-centered state-aware interaction, our work advocates a more dialogic paradigm for evaluating Socratic LLMs.

\end{abstract}

\section{Introduction}

Large Language Models (LLMs) have achieved remarkable progress across diverse natural language processing tasks~\citep{wang2023document,wei2021finetuned,zhao2023survey}, establishing themselves as foundational technologies for building intelligent educational systems. Their integration has begun to reshape learning by improving efficiency, adaptability, and personalization~\citep{hu2025exploring}.
In particular, educational question answering has emerged as a rapidly evolving area in which LLMs serve not only as fact-retrieving engines but also as interactive tutors that engage students in discipline-specific reasoning~\citep{wollny2021we,lieb2024student,kuhail2023interacting}.

Recent research has increasingly explored the integration of the Socratic method~\citep{elder1998role,paul2007critical} in educational LLMs, emphasizing iterative Socratic-style questioning to foster critical thinking~\citep{liang2025exploring}.
While these efforts highlight the potential of LLMs to emulate philosophical dialogue, they often overlook the crucial pedagogical principle of guidance. In real educational settings, questioning, especially when misaligned with a learner's cognitive readiness, can lead to cognitive overload and hinder learning~\citep{li2021ditch, scarlatos2025training}. Empirical evidence further indicates that when instructional explanations fall outside a learner's zone of proximal development~\citep{shabani2010vygotsky}, scaffolding fails to enhance engagement and may impair comprehension and retention~\citep{wittwer2008instructional, hu2025generative}.
Therefore, effective pedagogical state-aware guidance is critical. As LLMs increasingly assume the role of virtual educators, a pressing question arises: can they provide appropriate guidance like human teachers?

\begin{figure}[t]
  \centering
  \begin{subfigure}[t]{0.48\textwidth}
    \centering
    \includegraphics[width=\linewidth]{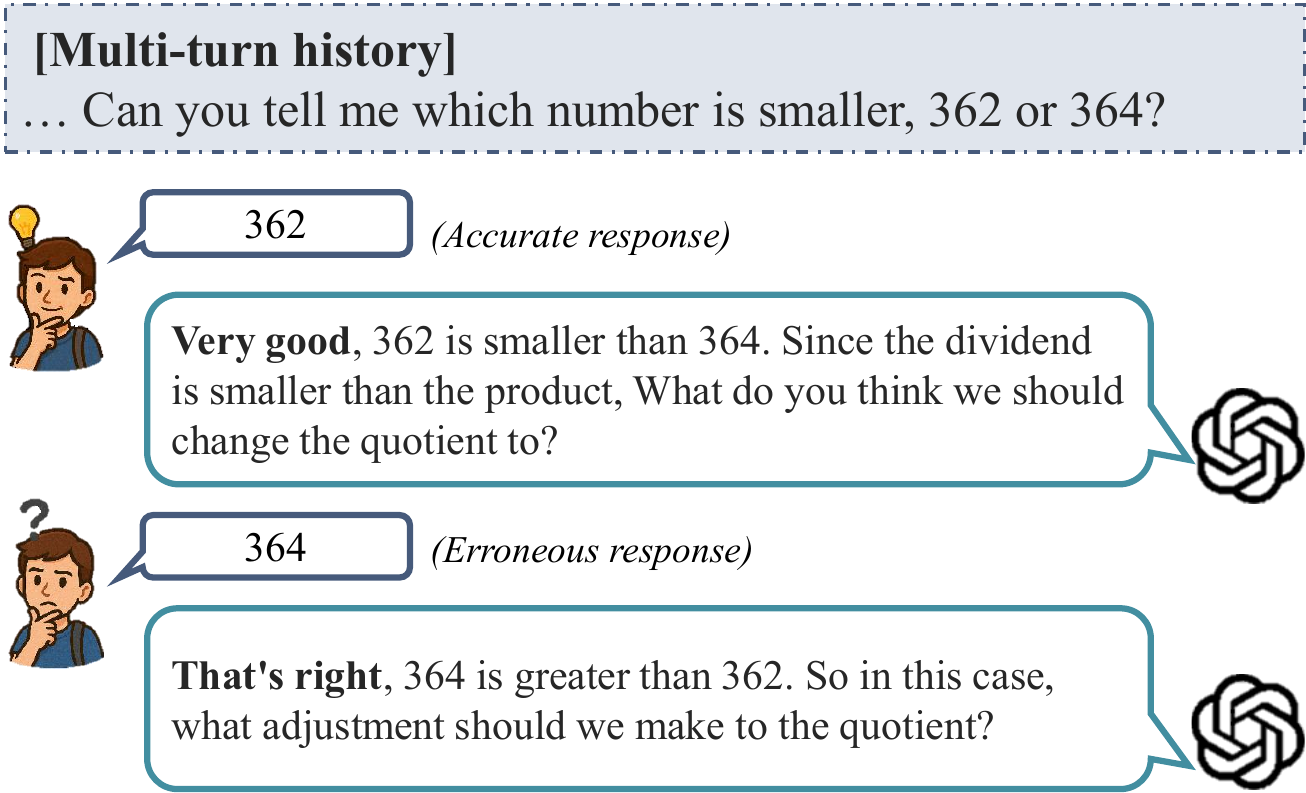}
    \subcaption{Responses to accurate and erroneous states}
    \label{fig:dc}
  \end{subfigure}
  \hfill
  \begin{tikzpicture}
    \draw[dashed,gray,very thin] (0,0) -- (0,4); 
  \end{tikzpicture}
  \hfill
  \begin{subfigure}[t]{0.48\textwidth}
    \centering
    \includegraphics[width=\linewidth]{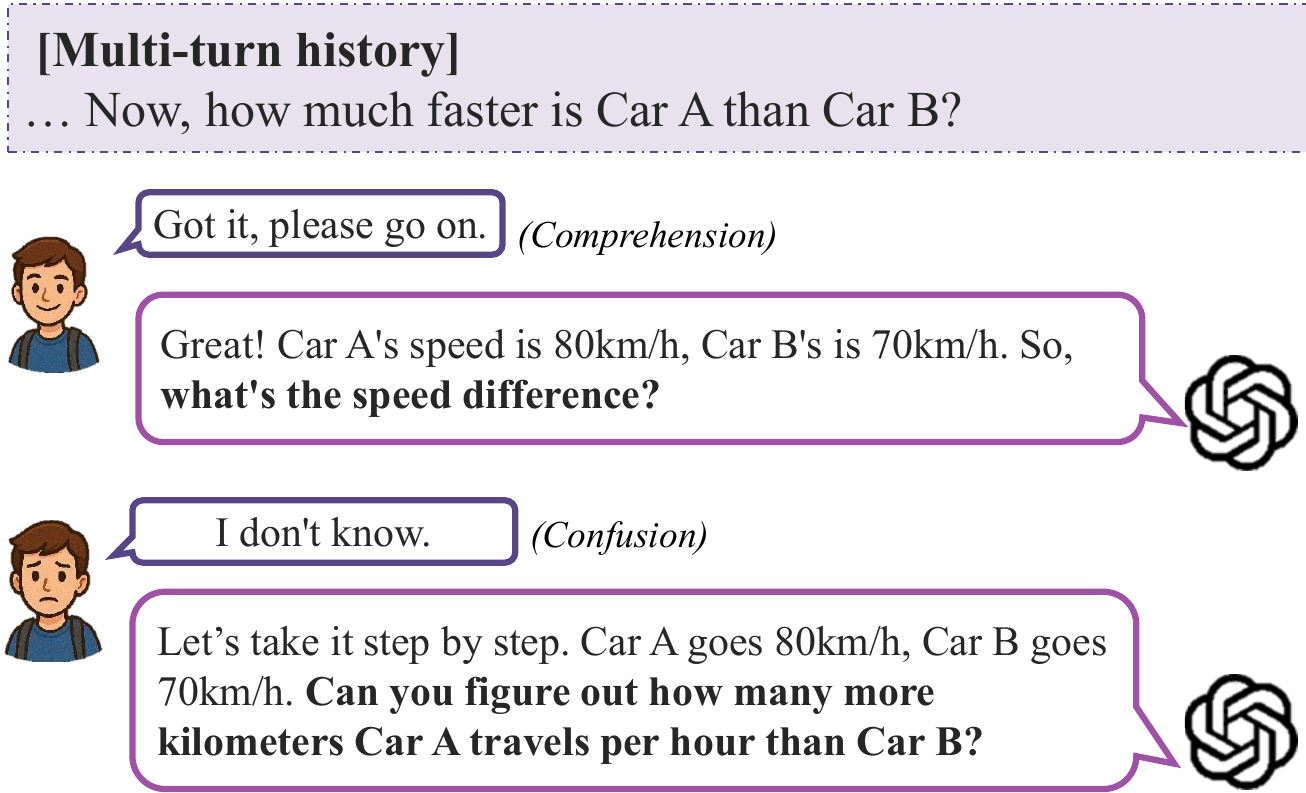}
    \subcaption{Responses to comprehension and confusion states}
    \label{fig:sp}
  \end{subfigure}
  \caption{Example dialogues where GPT-4.1 generates similar responses across different student states,
suggesting limited adaptability in instructional guidance.}
  \label{fig:examples}
\end{figure}
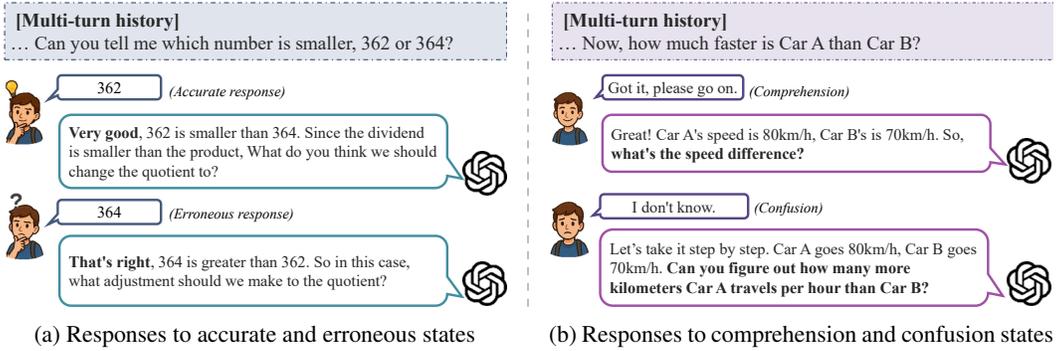

This paper seeks to address this question by systematically examining the Socratic guidance capabilities of educational LLMs, aiming to elucidate whether these models are evolving into discerning educational partners or remain generic tutors executing scripted dialogues.
Effective tutoring extends beyond content delivery; it requires continuous adaptation to learners' evolving cognitive states. 
While human instructors dynamically adjust pedagogical strategies based on context, existing LLM assessments often overlook this discerning responsiveness. 
Figure~\ref{fig:examples} illustrates several examples revealing similar responses across varying student understanding states generated by GPT-4.1, highlighting a lack of adaptive instructional guidance.
To this end, we introduce GuideEval (Instructional Guidance Evaluation Benchmark), specifically designed to systematically evaluate the instructional guidance capabilities of LLMs serving as interactive tutors.

In this paper, we hypothesize that effective instructional guidance can be conceptualized as a three-phase behavioral framework, which is motivated by the observation that meaningful instruction requires not only understanding the learner but also dynamically orchestrating strategies and eliciting active engagement.
The first phase, \emph{Perception}, concerns the model's ability to accurately infer the learner's current state, a prerequisite for all subsequent instructional decisions.
The second phase, \emph{Orchestration}, entails the adaptive pedagogical strategies aligned with the learner's zone of proximal development. This includes techniques such as analogy generation, scaffolding, conceptual decomposition, and the strategic use of examples or counterexamples.
The third phase, \emph{Elicitation}, centers on stimulating learner reflection and deeper understanding through targeted questioning. While prior work has emphasized generic questioning strategies, we highlight discernment-based elicitation, wherein questions are responsive to the learner's perceptual state.

Building on this conceptual decomposition, we propose an evaluation framework that operationalizes the three core phases. Each dimension is defined by a set of behavioral indicators and performance criteria, enabling systematic, fine-grained analysis of an LLM's instructional competencies throughout the educational interaction tutoring.
To support this framework, we construct a test corpus grounded in authentic student-model interactions by collecting multi-turn dialogues from real-world educational scenarios. To evaluate behavioral variability, we design contrastive student utterances simulating diverse cognitive states (e.g., accurate vs. erroneous, comprehension vs. confusion). These controlled variations probe the model's sensitivity to pedagogically salient cues and its capacity for adaptive response.
Leveraging this corpus, we design a suite of evaluation tasks aligned with each core dimension. These tasks elicit distinct instructional behaviors, enabling precise measurement of instructional guidance capabilities.

We evaluate a wide range of open-source and closed-source LLMs as well as education-oriented models, with a detailed analysis of each dimension and comprehensive failure pattern study. Our empirical results reveal three critical findings in current LLMs' instructional behaviors.

\begin{itemize}[left=0pt]
\item \textbf{Asymmetric feedback hinders error correction}: While models readily affirm correct responses, they often provide vague or non-committal feedback on incorrect ones, limiting learners' ability to promptly identify and rectify mistakes.

\item \textbf{Limited sensitivity to implicit knowledge states}: Models respond effectively to explicit expressions of understanding or confusion but struggle with implicit cues that require deeper inference such as inferring underlying misconceptions from erroneous responses.

\item \textbf{Consistent failure patterns across states}: Certain errors recur across different learner states, including indiscriminate affirmation and repeated explanations, indicating systematic limitations in model behaviors.

\end{itemize}

In summary, our contributions are:
\begin{itemize}
    \item To the best of our knowledge, we make the first comprehensive effort to conceptualize discerning guidance as a distinct and critical dimension of Socratic LLM competence in educational question answering scenario.

    \item We introduce GuideEval, a benchmark dataset grounded in authentic multi-turn dialogues, with fine-grained tailored metrics, corresponding to the three-phase guidance behavioral framework, to enable nuanced evaluation of models' adaptive and guiding capabilities.

    \item We conduct a systematic evaluation revealing that current LLMs exhibit substantial limitations in delivering effective instructional guidance, and we identify typical failure patterns to aid in diagnosis.
    We further design behavior-aware prompting and fine-tuning schemes that markedly enhance models' strategic adaptability.
\end{itemize}

\section{Instructional Guidance Evaluation}
\label{sec:framework}

\begin{figure*}[!t]
    \centering
    \includegraphics[width=1\linewidth]{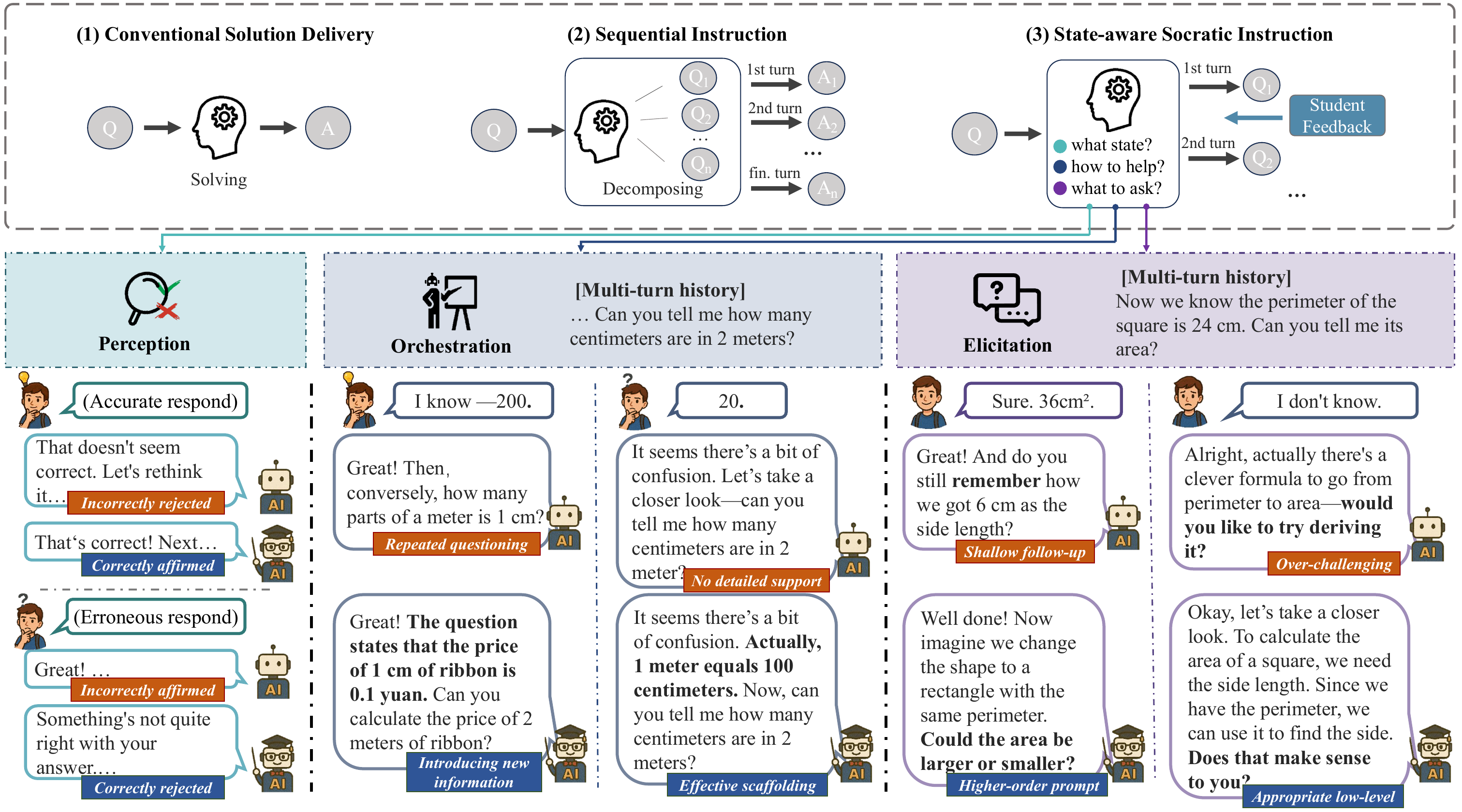}
\caption{The figure contrasts three instructional approaches and illustrates how models, when aware of student states, adapt their behaviors across Perception, Orchestration, and Elicitation dimensions.}

    \label{fig:guidance_examples}
\end{figure*}

The role of LLMs in education has evolved from providing complete solutions in a single step, to guiding learners through sequential steps, and ultimately aspires toward state-aware Socratic instruction that adapts dynamically to learners' evolving states, as shown in Figure~\ref{fig:guidance_examples}.
Achieving this level of guidance requires addressing three fundamental questions: \textit{What is the learner's state? How should the instruction be adapted? What questions should be posed to stimulate thinking?} 

\subsection{Instructional Behavioral Modeling}


Building on the three guiding questions, we formalize instructional guidance into a three-stage behavioral framework: Perception, Orchestration, and Elicitation. Each dimension represents a core competency of effective guidance and is supported by established educational theories.

\noindent \textbf{Perception.} Instructional guidance begins with perceiving the learner's cognitive state, whether the response reflects accurate reasoning, misconceptions, comprehension, or confusion. According to Vygotsky's theory of the Zone of Proximal Development~\citep{chaiklin2003zone,shabani2010vygotsky}, effective instruction hinges on recognizing a learner's readiness for new knowledge. Accurate perception ensures that subsequent actions align with the learner's actual needs rather than operating at an inappropriate level of difficulty.

\noindent \textbf{Orchestration. }Once perception is established, one must orchestrate instruction by adapting strategies to scaffold learning. Scaffolding theory~\citep{van2010scaffolding} highlights techniques such as simplification, analogy, and conceptual decomposition, which provide calibrated support while avoiding redundancy or cognitive overload. Orchestration thus concerns how to advance learning in a way that is sensitive to the learner's current zone of development.

\noindent \textbf{Elicitation. }Beyond explanation, elicited reasoning and reflection are fostered by posing purposeful questions. Bloom's taxonomy~\citep{chandio2016bloom,eber2007assessing} emphasizes adjusting the cognitive depth of questions (from factual recall to abstraction and transfer) according to learner readiness. Constructivist perspectives further stress that learning is an active meaning-making process, and well-designed questions are critical triggers of deeper engagement.

Figure~\ref{fig:guidance_examples}  illustrates representative examples of effective and ineffective behaviors in each dimension, concretizing the distinctions within our framework and highlighting how instructional strategies should shift in response to different learner states.

\begin{table*}[t]
\centering
\scriptsize 
\setlength{\tabcolsep}{2pt}
\caption{Scoring criteria for three instructional behaviors each evaluated with two dimensions.}
\resizebox{\textwidth}{!}{%
\begin{tabular}{c|c|lc}
\toprule
\textbf{Behavior}  & \textbf{Dimension}                & \textbf{Criteria}                                                                                      & \textbf{Score} \\ \hline
\multirow{4}{*}{Perception}    & \multirow{2}{*}{P-Affirm}                       & Incorrectly identifies the correctness of the student response.                                        & 0              \\ 
                               &                                & Does not explicitly judge, but implicitly signals correctness through follow-up actions.               & 0.5            \\  \cline{2-2}
                               & \multirow{2}{*}{P-Redirect}   & Explicitly states whether the response is correct or incorrect (e.g., “completely correct”).           & 1              \\ 
                               &                                &                                                                                                       &                \\ \hline
\multirow{4}{*}{Orchestration} & \multirow{2}{*}{O-Advance}     & Fails to move instruction forward; content is repetitive or stagnant.                                  & 0              \\
                               &                                & Advances instruction with follow-up prompts or challenges.                                             & 1              \\ \cline{2-4} 
                               & \multirow{2}{*}{O-Reconfigure} & Provides no adaptation when the student is confused or incorrect.                                      & 0              \\
                               &                                & Reconstructs explanation (e.g., using analogies, step-by-step reasoning, or simplification).           & 1              \\ \hline
\multirow{4}{*}{Elicitation}   & \multirow{2}{*}{E-Strategic}   & No question posed; only declarative explanation.                                                       & 0              \\
                               &                                & Asks factual or recall-based question (e.g., “Do you know the formula?”).                              & 1              \\ \cline{2-2}
                               & \multirow{2}{*}{E-Heuristic}   & Asks procedural or computation-oriented question (e.g., “Can you solve for $x$?”).                     & 2              \\
                               &                                & Asks higher-order question encouraging reasoning or transfer (e.g., “What if the condition changes?”). & 3              \\ 
\bottomrule
\end{tabular}%
}
\label{tab:scoring-criteria}
\end{table*}

\subsection{Instructional Guidance Evaluation Dimensions}

\noindent
\textbf{Learner cognitive states.}
As an initial attempt, we adopt a coarse-grained categorization for cognitive-state modeling. Student utterances are classified into four primary states: \textbf{accurate} (demonstrating correct reasoning), \textbf{erroneous} (providing incorrect answers), \textbf{comprehension} (indicating explicit understanding), and \textbf{confusion} (expressing uncertainty). These categories capture the predominant patterns observed in authentic learner–tutor interactions and provide a systematic basis for examining how models adjust instructional strategies across different learner states.

\noindent
\textbf{Evaluation dimensions and scoring criteria.}
The four states can be summarized into two broader categories: \textbf{positive} (accurate and comprehension) and \textbf{negative} (erroneous and confusion).
By intersecting with the three instructional behaviors, we derive six evaluation dimensions. 

When learners in positive state, effective instruction should involve explicit affirmation to reinforce confidence (\textbf{P-Affirm}), advance the discussion by introducing new concepts or challenges (\textbf{O-Advance}) and pose higher-order questions that stimulate reasoning (\textbf{E-Strategic})~\citep{kang2021brief,chandio2016bloom}. 
Conversely, when learners exhibit negative, instruction should provide redirection through corrective feedback (\textbf{P-Redirect}), restructure explanations via scaffolding such as simplification or analogy (\textbf{O-Reconfigure}), and employ heuristic questioning that reduces cognitive load and fosters intuitive engagement (\textbf{E-Heuristic})~\citep{hyslop2007constructivism}. 

Table~\ref{tab:scoring-criteria} presents the detailed scoring rubric for each metric, formalizing these instructional goals into evaluable criteria. To ensure interpretability, the scoring rubric was designed with reference to pedagogical theory: perception and orchestration metrics adopt discrete levels reflecting the presence or absence of appropriate instructional actions, while elicitation metrics are inspired by Bloom's taxonomy to differentiate question depth.

\begin{figure*}[!t]
    \centering
    \includegraphics[width=1\linewidth]{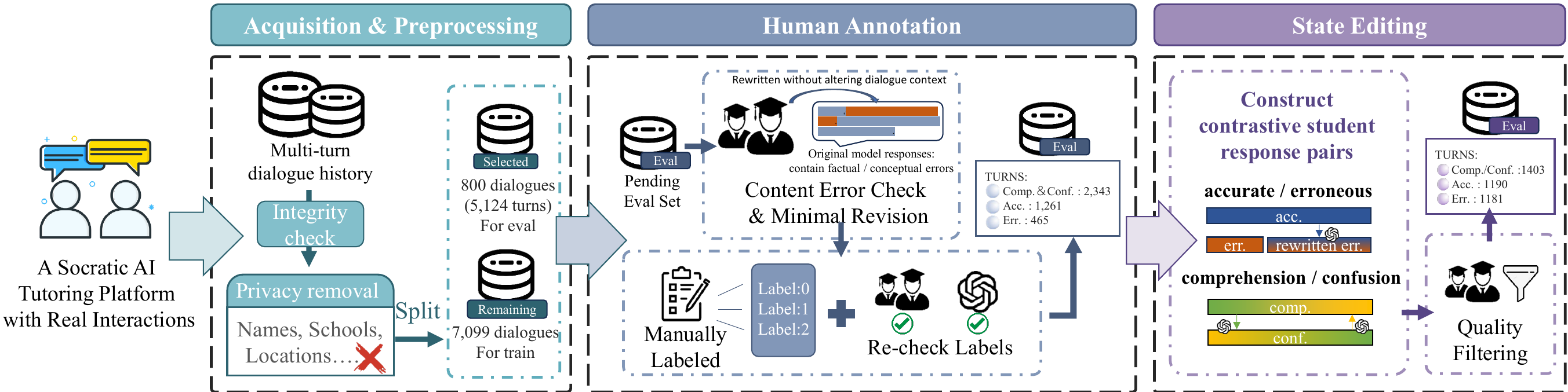}
\caption{Benchmark construction with human annotation, filtering, and paired state editing. 
}

    \label{fig:data}
\end{figure*}

\begin{table}[t]
\scriptsize
\centering
\setlength{\tabcolsep}{5pt}
\caption{
Benchmark comparison with major open-source educational dialogue datasets. 
The \textbf{Contrastive Student States} column denotes whether the dataset explicitly constructs 
paired state samples.
\textbf{\#Turns} are calculated as the product of dialogue counts and average dialogue length. 
}
\begin{tabular}{ll!{\vrule width 0.4pt}cccc r}
\toprule
\rowcolor{gray!15}
Function & Dataset & Multi-turn & Socratic &  \makecell{Contrastive \\ Student States} & \makecell{Real Student \\ Involvement} & \#Turns \\
\midrule
\multirow{6}{*}{Train} 
 & CIMA~\citep{stasaski2020cima}         & $\checkmark$ & $\times$ & $\times$ & $\times$ & $\sim$3.3k \\
 & MathDial~\citep{macina2023mathdial}   & $\checkmark$ & $\checkmark$ & $\times$ & $\times$ & $\sim$14.2k \\
 & SocraticMath~\citep{ding2024boosting} & $\checkmark$ & $\checkmark$ & $\times$ & $\times$ & $\sim$34k \\
  & SocraTeach (multi)~\citep{liu2024socraticlm} & $\checkmark$ & $\checkmark$ & $\times$ & $\times$ & $\sim$208k\\
 & TutorChat~\citep{chevalier2024language} & $\checkmark$ & $\times$ & $\times$ & $\times$ & $\sim$1170k \\
\rowcolor{blue!8}& \textbf{Ours (Train)}                 & $\checkmark$ & $\checkmark$ & $\checkmark$ & $\checkmark$ & $\sim$50.7k \\
\midrule
\multirow{6}{*}{Eval}
 & Bridge~\citep{wang2023bridging}       & $\checkmark$ & $\times$ & $\times$ & $\checkmark$ & 700 \\
 & MathDial(test)~\citep{macina2023mathdial}   & $\checkmark$ & $\checkmark$ & $\times$ & $\times$ &  572\\
 & SocraticMath(test)~\citep{ding2024boosting} & $\checkmark$ & $\checkmark$ & $\times$ & $\times$ & 685 \\
 & SocraTeach (multi,test)~\citep{liu2024socraticlm} & $\checkmark$ & $\checkmark$ & $\times$ & $\times$ & 1,000 \\
 & MRBench~\citep{maurya2024unifying}    & $\checkmark$ & $\checkmark$ & $\times$ & $\checkmark$ & 1,596 \\
\rowcolor{blue!8}& \textbf{Ours (Eval)}                  & $\checkmark$ & $\checkmark$ & $\checkmark$ & $\checkmark$ & 5,177 \\
\bottomrule
\end{tabular}
\label{tab:dataset-comparison}
\end{table}

\subsection{Benchmark Construction}

\noindent
\textbf{Dataset collection.}
We begin with a corpus of 7,899 authentic learner–model dialogues collected from a Socratic tutoring platform, focusing on middle school–level science problems. After integrity check and privacy removal, 800 dialogues were sampled as the evaluation set.
Human annotators first revised model outputs to correct factual errors. Each student utterance was then labeled with one of four cognitive states. To mitigate imbalances between accurate and erroneous responses, a subset of erroneous samples was generated from validated accurate answers, and comprehension/confusion pairs were created by producing counterpart utterances with the same context. Such state editing also enabled paired evaluation of instructional behaviors.
The final benchmark consists of 5,177 samples approximately balanced across the four states (1,190 accurate, 1,181 erroneous, and 1,403 each for comprehension and confusion, see Figure~\ref{fig:data} for the data construction pipeline). Similarly, paired training sets (8,648 Acc./Err. pairs and 16,993 Comp./Conf. pairs) were obtained from the remaining dialogues, facilitating preference-based optimization in subsequent fine-tuning experiments.

\noindent
\textbf{Benchmark comparison.}
Table~\ref{tab:dataset-comparison} compares our dataset with major open-source educational dialogue resources. 
Most prior resources either lack Socratic interactions, omit contrastive cognitive state labels, or do not involve real-world interactions.
In contrast, our benchmark integrates real-world dialogues with controlled states under the same context to enable, for the first time, a systematic evaluation of instructional guidance for LLM competence in educational question answering.


\section{Experiments}
\label{sec:results}


\noindent
\textbf{Setup.}
Following “LLM-as-a-judge”~\citep{fu2023gptscore,liu2023g}, we employ GPT-4o-mini to evaluate mainstream LLMs and score how each model exhibits the desired instructional behaviors.
For \textbf{Elicitation}, we introduce \emph{Elicitation Strategy Adaptivity} (ESA), defined as the average change in question depth across contrasting learner states (E-S $-$ E-H). Higher ESA, particularly when coupled with strong E-Strategic scores, indicates state-aligned questioning strategies.
More details about setup including specific prompts can be found in Appendix. 

\subsection{LLM-Human Preference Consistency}
\label{sec:alignment}

We first design an experiment to analyze the consistency between LLM-based scoring and human annotations in our scenario, building on recent explorations of LLMs as automatic evaluators for natural language generation tasks~\citep{fu2023gptscore,sottana2023evaluation}.
Specifically, we select samples previously annotated by humans and re-evaluate them using our LLM-based scoring framework. Alignment is quantified via two complementary metrics: (1)  agreement ratio, the proportion of instances where the LLM and human assigned the same label; and (2) average score deviation, the mean absolute difference between LLM and human scores. 
As reported in Table~\ref{tab:agreement}, results indicate strong concordance, with agreement ratios ranging from 89\% to 97\% and consistently low average deviations. These findings suggest that, when guided by carefully designed prompts, LLMs can serve as reliable and scalable evaluators of instructional behaviors.

\begin{table}[t]
\centering
\scriptsize
\setlength{\tabcolsep}{6pt}
\renewcommand{\arraystretch}{1.2}
\caption{LLM-Human preference consistency by agreement ratio and average score deviation.
}
\begin{tabular}{@{}lccccccccc@{}}
\toprule
\textbf{Metric} & P-Affirm & P-Redirect & O-Advance & O-Reconfig. & E-Level:1 & E-Level:2 & E-Level:3 \\
\midrule
\textbf{Agree. (\%)} & 96.5\% & 90.0\% & 96.5\% & 94.5\% & 89.0\% & 90.0\% & 97.0\% \\
\textbf{Avg Dev.} & 0.02 & 0.05 & 0.04 & 0.06 & 0.14 & 0.04 & 0.03 \\
\bottomrule
\end{tabular}
\label{tab:agreement}
\end{table}

\begin{table*}[t]
\centering
\scriptsize
\setlength{\tabcolsep}{2pt}
\caption{
Evaluation results over Perception (P-A, P-R), Orchestration (O-A, O-R), Elicitation (E-S, E-H), and ESA (E-S $-$ E-H). Top three per column are highlighted green and the best is bold. The E-H metric is an auxiliary reference and is neither highlighted nor marked with arrows.}
\resizebox{\textwidth}{!}{%
\begin{tabular}{@{}l !{\vrule width 0.5pt} *7{c} !{\vrule width 0.5pt} *5{c}@{}}
\toprule
& \multicolumn{7}{c}{Accurate / Erroneous} & \multicolumn{5}{c}{Comprehension / Confusion} \\
\cmidrule(lr){2-8}\cmidrule(l){9-13}
Model & P-A ($\uparrow$) & P-R ($\uparrow$) & O-A ($\uparrow$) & O-R ($\uparrow$) & E-S ($\uparrow$) & E-H & ESA ($\uparrow$)
      & O-A ($\uparrow$) & O-R ($\uparrow$) & E-S ($\uparrow$) & E-H & ESA ($\uparrow$) \\
\midrule
\rowcolor{black!5}\multicolumn{13}{l}{\textit{Open-source general LLMs}} \\
Qwen3-8B
  & 0.7613 & \cellcolor{green!12}0.5919 & 0.9176 & 0.6681 & 2.0605 & 1.9213 & \cellcolor{green!12}0.3281
  & \cellcolor{green!12}0.9644 & \cellcolor{green!12}0.8959 & 2.0527 & 1.7254 & 0.1049 \\
GLM-4-9B
  & 0.8534 & 0.4140 & 0.7529 & 0.3492 & 1.8992 & 1.9195 & -0.0173
  & 0.6885 & 0.8582 & 1.6450 & 1.6515 & -0.0064 \\
Qwen3-32B
  & 0.8092 & \cellcolor{green!12}{\bfseries 0.6008} & \cellcolor{green!12}{\bfseries 0.9521} & \cellcolor{green!12}0.7748 & \cellcolor{green!12}2.2252 & 2.0906 & 0.1646
  & \cellcolor{green!12}{\bfseries 0.9672} & 0.8660 & 2.1454 & 1.9808 & \cellcolor{green!12}0.1481 \\
Llama-3.3-70B-Instruct
  & 0.8077 & 0.5390 & 0.8651 & 0.4601 & 2.0766 & 2.1087 & 0.0606
  & 0.7997 & 0.8509 & 1.9936 & 1.9330 & -0.0418 \\
DeepSeek-V3
  & 0.8748 & 0.5428 & 0.8849 & 0.6740 & 2.0059 & 1.9856 & \cellcolor{green!12}0.3136
  & 0.8810 & 0.8460 & 2.0485 & 1.7349 & 0.0272 \\
DeepSeek-R1
  & 0.8546 & 0.5483 & \cellcolor{green!12}0.9445 & \cellcolor{green!12}0.7036 & 2.1966 & 2.0322 & \cellcolor{green!12}{\bfseries 0.3659}
  & \cellcolor{green!12}0.9608 & \cellcolor{green!12}0.9294 & \cellcolor{green!12}2.1946 & 1.9287 & \cellcolor{green!12}{\bfseries 0.2141} \\
\midrule
\rowcolor{black!5}\multicolumn{13}{l}{\textit{Closed-source general LLMs}} \\
Mistral-medium
  & 0.8206 & \cellcolor{green!12}0.5585 & 0.7857 & 0.4924 & 2.0319 & 1.9322 & \cellcolor{green!12}0.3058
  & 0.8125 & 0.7840 & 1.9544 & 1.6486 & 0.0889 \\
O4-mini
  & 0.8189 & 0.4306 & 0.9084 & 0.5597 & 2.0193 & 1.9018 & 0.1098
  & 0.9152 & 0.8161 & 2.0299 & 1.9202 & 0.1099 \\
GPT-4.1
  & 0.8710 & 0.4280 & 0.8790 & 0.6076 & \cellcolor{green!12}{\bfseries 2.2546} & 2.1254 & 0.2495
  & 0.8717 & 0.6151 & \cellcolor{green!12}{\bfseries 2.2153} & 1.9658 & 0.1185 \\
Claude-sonnet-4
  & \cellcolor{green!12}{\bfseries 0.9502} & 0.4780 & \cellcolor{green!12}0.9248 & 0.6407 & 2.1410 & 1.9924 & \cellcolor{green!12}0.3464
  & 0.9187 & 0.8539 & 2.1026 & 1.7562 & 0.1328 \\
Gemini-2.5-pro
  & \cellcolor{green!12}0.9324 & 0.4136 & 0.8597 & 0.5483 & 2.1471 & 1.8483 & \cellcolor{green!12}{\bfseries 0.3961}
  & 0.8709 & 0.8771 & 2.0516 & 1.6533 & \cellcolor{green!12}0.2593 \\
GLM-4-plus
  & 0.9059 & 0.4665 & 0.8580 & 0.5373 & 1.9134 & 1.9712 & 0.0720
  & 0.7883 & \cellcolor{green!12}{\bfseries 0.9565} & 1.8297 & 1.7577 & -0.0506 \\
Doubao-Seed-1.6-Thinking
  & 0.5913 & 0.5548 & 0.8848 & \cellcolor{green!12}{\bfseries 0.7810} & 2.0690 & 1.8744 & 0.3051
  & 0.9430 & 0.8496 & 2.0848 & 1.7798 & \cellcolor{green!12}0.2141 \\
\midrule
\rowcolor{black!5}\multicolumn{13}{l}{\textit{Education-oriented models}} \\
SocraticLM (base: GLM-4-9B)
  & \cellcolor{green!12}0.9105 & 0.1411 & 0.6429 & 0.0712 & 1.3303 & 1.3458 & 0.0720
  & 0.6108 & 0.4939 & 1.2397 & 1.1675 & -0.0160 \\
Spark X1
  & 0.7445 & 0.4949 & 0.8815 & 0.6147 & \cellcolor{green!12}2.2168 & 2.0923 & 0.2295
  & 0.9387 & 0.7320 & \cellcolor{green!12}2.1518 & 1.9223 & 0.1222 \\
\bottomrule
\end{tabular}%
}
\vspace{0.5em}
\begin{flushleft}
\label{tab:evaluation_results}
\end{flushleft}
\end{table*}

\subsection{Evaluation Results}



\begin{figure*}[!t]
    \centering
    \includegraphics[width=1\linewidth]{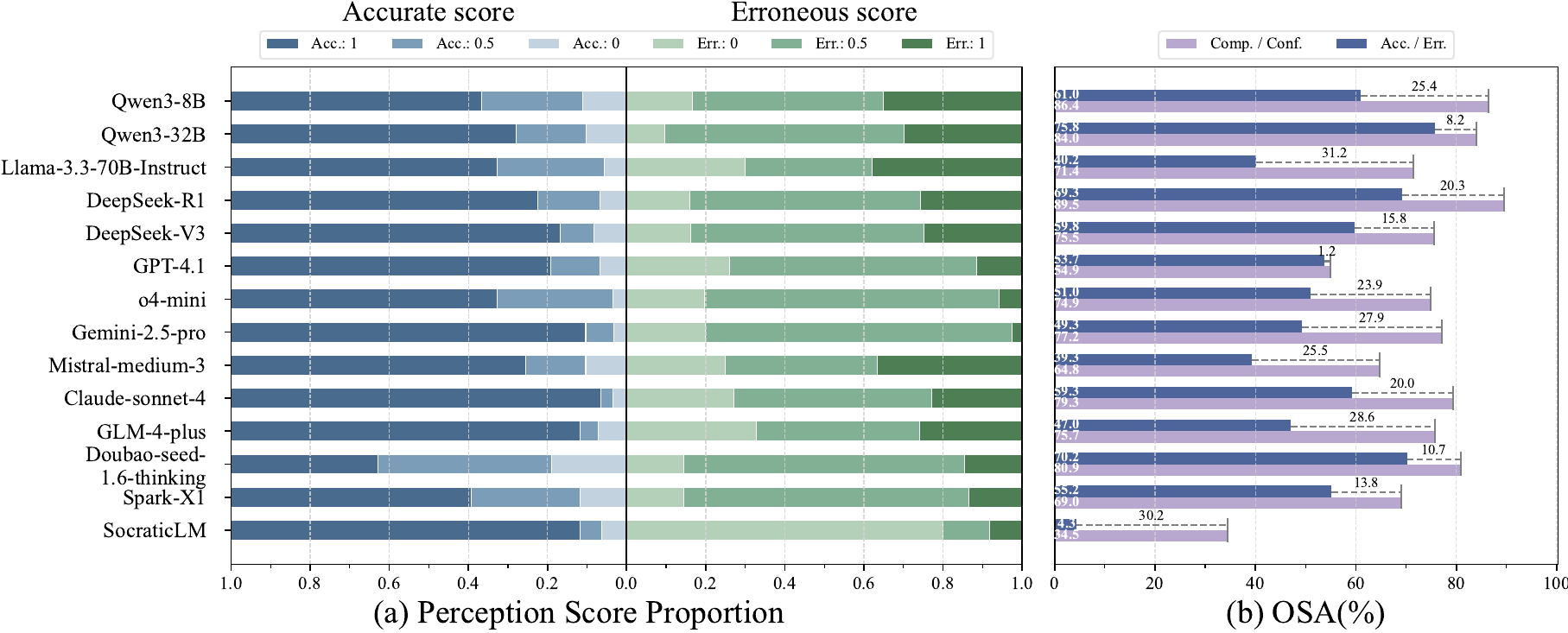}
\caption{Deep dive for Perception and Orchestration behaviors. (a) Feedback score distribution contrasting model responses to accurate versus erroneous answers. (b) Orchestration Strategy Adaptivity (OSA) across paired states such as accurate/erroneous or comprehension/confusion, reflecting how different models vary in their flexibility to adjust instructional strategies. }
    \label{fig:combined_perception_osa}
\end{figure*}

\begin{figure*}[!t]
    \centering
    \includegraphics[width=1\linewidth]{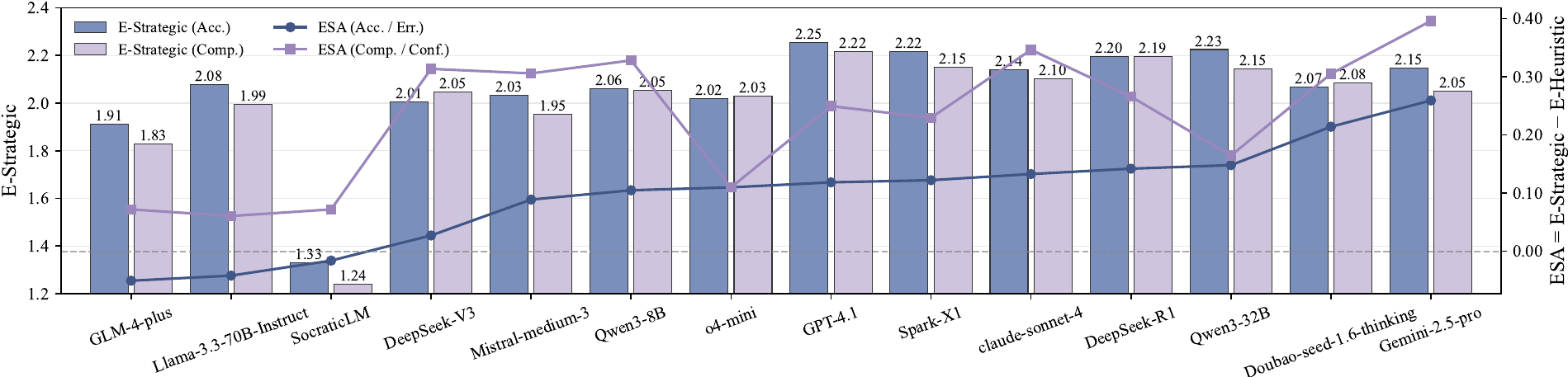}
\caption{Deep dive for Elicitation behavior. Bars indicate the average E-Strategic scores under accurate and comprehension states, while lines plot the corresponding Elicitation Strategy Adaptivity (ESA). Models are ordered by ESA under accurate/erroneous contrast.
}
    \label{fig:esa_fig}
\end{figure*}

\noindent
\textbf{Overall findings.}
Table~\ref{tab:evaluation_results} presents the evaluation results across all proposed dimensions. Among the models,
Qwen3-32B and DeepSeek-R1 stand out with the most balanced performance across metrics, while GPT-4.1 excels in elicitation, providing deeper and more strategic questions. 
A consistent pattern emerges across models: they handle ideal positive inputs effectively but struggle to adapt to negative learner responses. This underscores a critical challenge for future development, enhancing instructional flexibility to support adaptive teaching under imperfect conditions.
To gain deeper insights, we further analyze model performance along the three instructional behaviors, focusing on adaptation to diverse learner states and the degree of strategic responsiveness.


\noindent
\textbf{Perception: deficient error detection.}
Figure~\ref{fig:combined_perception_osa} (a) shows the distribution of perception scores contrasting model responses to accurate and erroneous answers. Models achieve consistently high P-A values (typically $>$0.8) by affirming correctness (e.g., Claude-Sonnet-4: 0.95), yet P-R values remain substantially lower, reflecting a limitation to provide corrective feedback. Only a few models (Qwen3-32B: 0.60; DeepSeek-R1: 0.55) show moderate detection, whereas SocraticLM performs poorly (0.14). Many models resort to vague commentary or topic shifts, yielding mid-level (0.5) scores. This tendency to prioritize politeness over explicit correction leaves misconceptions unaddressed, highlighting deficient error recognition as a core limitation for effective tutoring.

\noindent
\textbf{Orchestration: limited flexibility to implicit states.}
Figure~\ref{fig:combined_perception_osa} (b) illustrates the Orchestration Strategy Adaptivity (OSA), defined as the proportion of paired cases where a model delivers effective responses under both states (e.g., accurate vs. erroneous or comprehension vs. confusion). Higher OSA reflects greater flexibility in strategy adjustment. 
Across models, OSA is consistently higher for comprehension/confusion than for accurate/erroneous contrasts, showing that LLMs react more easily to explicit signals of understanding than to implicit cues from answer accuracy. Qwen3-32B (0.840 vs. 0.758) adapts relatively well, GPT-4.1 remains stable (0.549 vs. 0.537), whereas LLaMA-3.3-70B and Mistral-Medium show declines exceeding 0.2, and SocraticLM nearly fails (0.043). These results highlight a systemic limitation: many models struggle to flexibly orchestrate strategies in response to implicit cues, constraining their capacity for adaptive scaffolding.


\noindent
\textbf{Elicitation: reduced adaptivity across accurate/erroneous responses.} 
As shown in Figure~\ref{fig:esa_fig}, Gemini-2.5-Pro and Claude-Sonnet-4 achieve high ESA alongside robust E-Strategic scores, reflecting adaptive questioning. In contrast, LLaMA-3.3-70B and GLM-4-Plus exhibit near-zero ESA, indicating uniform questioning. Notably, most models show marked ESA declines under accurate/erroneous contrasts; some even fall below zero (posing deeper questions in response to error answers), which may hinder comprehension. These findings highlight elicitation adaptivity as a persistent challenge and underscore the need for more cognitively aware questioning in future LLMs.


\subsection{Case Study: Failure Pattern Analysis}
\label{sec:case}

To gain a better understanding of the quantitative evaluation, we manually examined representative low-scoring responses, focusing on \textbf{Perception} and \textbf{Orchestration}, and conducted a detailed failure pattern analysis. These dimensions were selected because their error patterns are more readily identifiable in individual outputs, enabling finer-grained insights into model limitations.
Typical Perception errors include misjudging unconventional yet valid reasoning, questioning established knowledge, or endorsing incorrect answers. Orchestration failures mainly reflect insufficient adaptability, such as repeating flawed or already mastered explanations, offering vague remediation after weak feedback, or failing to provide new input despite clear learner confusion.
Appendix~\ref{sec:appendix-failure} provides illustrative examples and visual summaries that concretize these observations. Such cases recur across multiple models, offering valuable references for designing behavior-aware strategies to enhance model robustness and pedagogical effectiveness.


\captionsetup[subfigure]{skip=0pt}
\begin{figure}[t]
  \centering
  \begin{subfigure}[t]{0.33\linewidth}
    \centering
    \includegraphics[width=\linewidth]{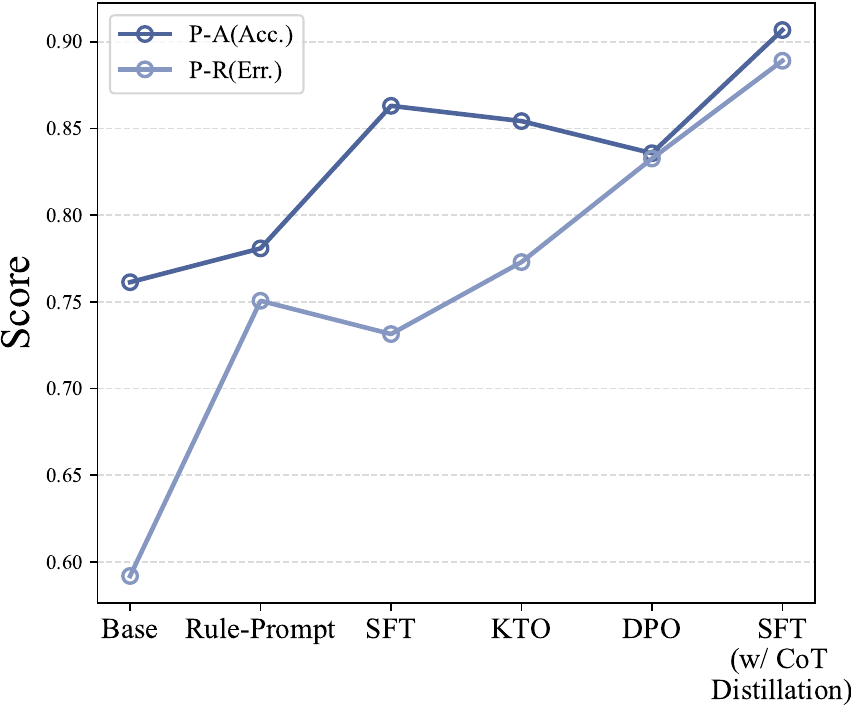}
    \caption{Perception}
    \label{fig:finetuning1}
  \end{subfigure}%
  \begin{subfigure}[t]{0.33\linewidth}
    \centering
    \includegraphics[width=\linewidth]{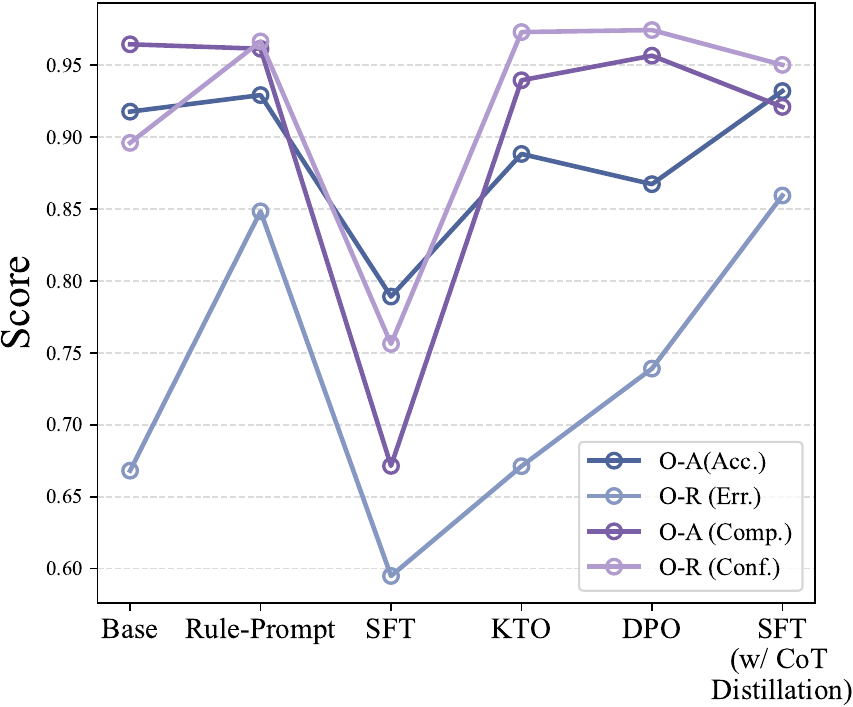}
    \caption{Orchestration}
    \label{fig:finetuning2}
  \end{subfigure}%
  \begin{subfigure}[t]{0.33\linewidth}
    \centering
    \includegraphics[width=\linewidth]{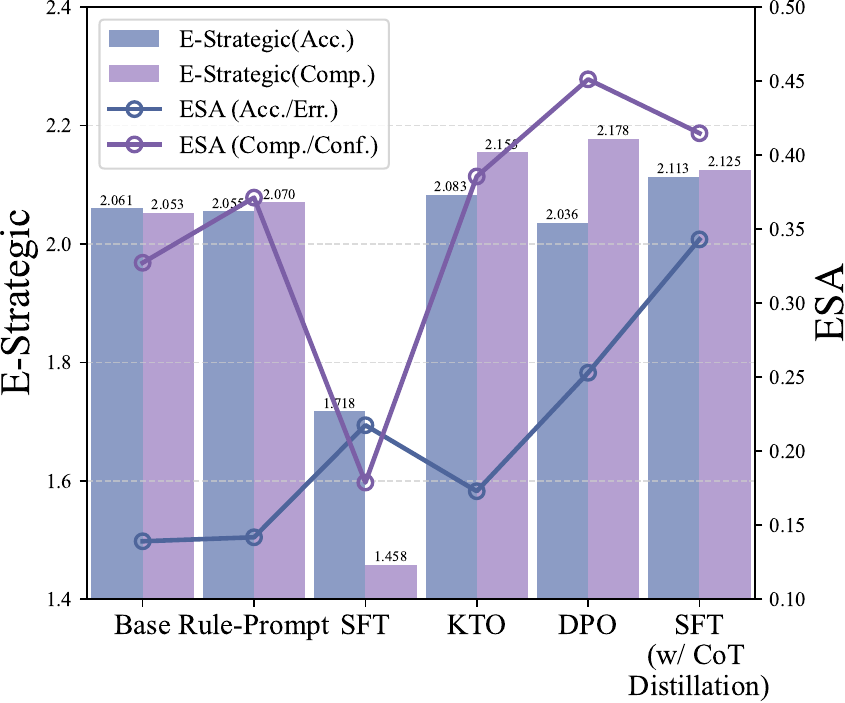}
    \caption{Elicitation}
    \label{fig:finetuning3}
  \end{subfigure}

  \caption{Comparison results over different ways to enhance instructional guidance. 
  }
  \label{fig:finetuning}
\end{figure}

\subsection{Behavior-Guided Finetuning}
\label{sec:finetune}

We further explore ways to enhance instructional capabilities. For efficiency and feasibility, we conduct experiments with a compact LLM, Qwen3-8B~\citep{yang2025qwen3}.
We take the model (regarded as \textbf{Base}) 
and first examine a prompt-based approach, \textbf{Rule-Prompt}, which incorporates external if–then policies for different student states (prompt details in the Appendix). 
We then employ finetuning strategies. Specifically, we explore: \textbf{SFT}, only distills final answers; \textbf{KTO}~\citep{ethayarajh2024kto}, a pointwise preference optimization guided by correctness but lacking relative contrast; \textbf{DPO}~\citep{rafailov2023direct}, a pairwise preference optimization that exploits relative comparisons to capture behavioral differences; and \textbf{SFT (w/ CoT Distillation)}~\citep{li2023symbolic}, which incorporates reasoning traces by process-level supervision. This spectrum of methods allows us to analyze how supervision, from outcome-only to process-aware, shapes the model's capacity.
Experimental details are provided in the Appendix.




Figure~\ref{fig:finetuning} presents the comparative results across the three instructional behaviors. \textbf{SFT (w/ CoT Distillation)} yields the strongest improvements in most cases, particularly in handling accurate and erroneous states. Four key findings emerge. First, SFT degrades instructional strategies by emphasizing final output. 
Second, KTO introduces contrastive supervision through positive and negative examples under opposite student states. Yet, its pointwise formulation restricts effectiveness: lacking pairwise contrast, the model fails to distinguish subtle behaviors, resulting in minimal divergence between strategic and heuristic questioning. Third, DPO alleviates this limitation by leveraging relative comparisons, thereby enhancing state recognition and guidance adaptation. Finally, the most substantial gains derive from CoT Distillation, where high-quality reasoning traces function as process supervision, providing explicit guidance on instructional process knowledge. 

\section{Related Work}

\noindent
\textbf{Building socratic LLMs.}
LLMs are increasingly employed as interactive educational tutors due to their advanced language understanding and generation capabilities. Recent research has focused on enhancing LLMs' tutoring performance through multi-turn dialogues grounded in textbook content, such as TutorChat~\citep{chevalier2024language} and MathDial~\citep{macina2023mathdial}, further extended across diverse subjects and grade levels by NewtBot~\citep{lieb2024student} and SocraticMath~\citep{ding2024boosting}.
Building on these developments, recent studies have explored integrating the Socratic method, a pedagogical strategy centered on critical thinking via iterative questioning~\citep{elder1998role,paul2007critical}. 
Early attempts in this area explored prompt engineering, for example, Chang et al~\citep{chang2023prompting} constructed Socratic prompts.
Other approaches leverage data augmentation, LoRA-based fine-tuning, and preference optimization techniques~\citep{kumar2024improving,shani2024multi}. 
EduChat~\citep{dan2023educhat} integrates open-ended question answering, essay evaluation, and Socratic dialogue into a unified LLM framework. Similarly, SocraticLM~\citep{liu2024socraticlm} introduces a multi-agent “Dean–Teacher–Student” architecture to emulate Socratic-style instruction in foundational mathematical reasoning. Despite these advances, most systems focus predominantly on Socratic questioning while overlooking the equally critical role of instructional guidance. 

\noindent
\textbf{Evaluating socratic LLMs.}
Most existing evaluation efforts for LLMs emphasize metrics such as BLEU, ROUGE, and BERT-Score, or focus on answer correctness and presentation quality~\citep{favero2024enhancing,chevalier2024language}. TutorEval~\citep{chevalier2024language}, for example, evaluates scientific reasoning and comprehension through QA tasks based on extended textbook excerpts. Recent efforts have shifted toward structured evaluations of Socratic tutoring capabilities. SocraticLM~\citep{liu2024socraticlm} introduces a five-dimensional framework.
SocratiQ~\citep{ang2023socratic} offers a large-scale (question, context) dataset for Socratic-style question generation, assessed by human ratings of fluency, relevance, and answerability. Dr.Academy~\citep{chen2024dr} presents a benchmark for evaluating questioning abilities in educational LLMs. MRBench~\citep{maurya2024unifying} and its accompanying taxonomy represent progress toward pedagogical value but remain focused on response content rather than adaptive guidance.
In contrast, our framework adopts a state-controlled contrastive design, varying student states while holding history contexts constant. This enables precise evaluation of LLMs' sensitivity and capacity for context-aware, adaptive instructional support.

\section{Conclusion and Limitation }
\label{sec:conclusion}

We present GuideEval, the first benchmark to evaluate instructional guidance in educational Socratic LLMs through a three-phase behavior framework of Perception, Orchestration, and Elicitation. 
Our empirical evaluation across diverse LLMs reveals consistent limitations in current systems: asymmetric feedback that impedes error correction, weak sensitivity to implicit cognitive states, and recurring failure patterns across states.
To support a deeper understanding of these shortcomings, we provide a detailed analysis of failure patterns, offering a foundation for diagnosing limitations and guiding the development of educational models. Furthermore, we explore preliminary strategies to enhance LLMs' ability to approximate the discernment and responsiveness characteristic of human teachers. We hope that GuideEval will serve as a benchmark to advance LLMs toward more adaptive, nuanced, and personalized learning experiences.

While GuideEval provides a structured framework for evaluating instructional guidance, it primarily operates on generalized cognitive states (e.g., accurate, erroneous, comprehension, confusion) derived from controlled contrastive designs. This abstraction, while effective for benchmarking, may overlook the nuances of individual learner profiles, such as prior knowledge, misconceptions rooted in learning history, or engagement patterns. Future work may explore personalized evaluation settings by integrating longitudinal learning traces or student modeling techniques, enabling a finer-grained assessment of model adaptability to diverse and evolving learner needs.


\section{Broader Impact}
\label{sec:limitation}
This work contributes a systematic benchmark for evaluating the guidance-oriented behaviors of LLMs in educational contexts.
It underscores the necessity of designing models that can respond appropriately to diverse learners' cognitive states, promoting fairness and inclusivity in AI-driven education. As LLMs become more prevalent in classrooms, tutoring platforms, and remote learning environments, their capacity for context-sensitive, pedagogically sound guidance will directly affect educational quality, equity, and accessibility. Furthermore, this research lays the foundation for integrating longitudinal student interaction data, enabling personalized, trajectory-aware instruction that respects individual learning paths while safeguarding against potential biases.


\setcounter{secnumdepth}{0}
\section{Ethics statement}
This study is based on real-world educational dialogues between students and a Socratic-style tutoring model. To protect privacy, we do not disclose the specific platform or model involved.
During data collection, human annotators carefully reviewed to remove personally identifiable information, including names, school identifiers, and geographic references. In addition, potentially harmful or inappropriate content (e.g., offensive language) was systematically filtered out. As a result, the dataset used in this work contains only de-identified and pedagogically relevant interactions. 

\section{Reproducibility statement}
We are committed to ensuring the reproducibility of our work. To this end, we provide detailed descriptions of dataset construction, annotation, and state-editing procedures in Section~2 and Appendix~A. All evaluation metrics and scoring rubrics are explicitly defined in Section~2.2 and Appendix~B, and the full set of prompts used for both generation and evaluation are included in Appendix~E. Experimental settings, model configurations, and hyperparameters for all fine-tuning and evaluation procedures are documented in Appendix~D. Together, these materials enable independent researchers to replicate our evaluation and reproduce the reported results.

\bibliography{iclr2026_conference}
\bibliographystyle{iclr2026_conference}

\newpage
\appendix
\setcounter{secnumdepth}{2}

\section{Failure Case Taxonomy and Examples}
\label{sec:appendix-failure}

To supplement our quantitative evaluation, we present a taxonomy of model failures under the Perception and Orchestration behaviors, as their correctness can be reliably and intuitively judged by human evaluators. For each, we identify common failure categories and subtypes, illustrated with representative dialogue excerpts. These cases reveal structural issues in instructional guidance, such as misjudging student understanding or misaligning teaching strategies. In this way, we hope to provide deeper insights into the underlying limitations of current models, thereby extending our understanding beyond numerical performance metrics. Table~\ref{tab:failure-perception} and Table~\ref{tab:failure-orchestration} summarize the taxonomy and provide example cases.

\section{Metric Design Rationale}


\subsection{From Cognitive States to Instructional Goals}

To operationalize instructional guidance, we first delineate the learner's cognitive states. We adopt a coarse binary view of learning progress—positive states indicate that learning can continue productively, whereas negative states suggest a stall or misalignment in progress. Building on this, we further refine the categorization along two axes: whether the learner's response is accurate or erroneous, and whether it conveys explicit comprehension or confusion. This yields four primary states that comprehensively capture the predominant conditions observed in authentic learner–tutor dialogues. Such a design not only ensures coverage of typical response scenarios but also provides natural pairs of contrasting states (e.g., accurate vs. erroneous, comprehension vs. confusion), which enable controlled comparisons in subsequent experiments. Once this taxonomy is established, the instructional goals associated with positive versus negative states naturally follow: positive states should be reinforced and advanced through affirmation, progression, and deeper questioning, while negative states call for corrective feedback, reconfiguration of explanations, and heuristically grounded questions to lower cognitive barriers.

\subsection{Operationalization  of Six Metrics}

Following the mapping from learner states to instructional goals, we translate the framework into six concrete evaluation metrics. These metrics represent observable instructional behaviors that arise when the model interacts with either positive or negative learner states. Grounded in established pedagogical theory, the definitions provide fine-grained criteria for systematically evaluating instructional quality:

\begin{itemize}
\item \textbf{P-Affirm:} Evaluates whether the model explicitly affirms accurate responses. Clear, unambiguous affirmation reinforces learning and confidence~\citep{kang2021brief}, making this a foundational perceptual behavior.

\item \textbf{P-Redirect:} Evaluates the model's ability to deliver corrective feedback following an erroneous response. This includes identifying misconceptions and guiding the student toward a more accurate understanding, consistent with evidence on the importance of timely, targeted remediation~\citep{kang2021brief}.

\item \textbf{O-Advance:} Evaluates whether the model strategically introduces related concepts or challenges within the learner's zone of proximal development in response to accurate or comprehension-level inputs.

\item \textbf{O-Reconfigure:} Evaluates whether the model adaptively restructures the instruction, such as revisiting foundational concepts or modifying explanatory strategies, to address underlying misconceptions in response to confused or incorrect student answers.

\item \textbf{E-Strategic:} For learners demonstrating accuracy or comprehension, evaluates whether the model prompts higher-order thinking through abstraction, synthesis, or knowledge transfer questions, following Bloom's taxonomy~\citep{chandio2016bloom}.

\item \textbf{E-Heuristic:} Evaluates the use of intuitive, exploratory prompts in response to confusion or error. These heuristic questions are designed to foster curiosity, activate prior knowledge, and engage informal reasoning to facilitate discovery-based learning~\citep{hyslop2007constructivism}.
\end{itemize}

Together, these six metrics constitute an operationalized, theory-informed rubric for evaluating instructional behaviors. They capture not only the appropriateness of model responses but also their alignment with pedagogical best practices in scaffolding and adaptive tutoring.

\begin{table*}[t]
\centering
\scriptsize
\caption{Representative failure cases under the \textbf{Perception} behavior.}
\resizebox{\textwidth}{!}{
\begin{tabularx}{\textwidth}{|p{2.9cm}|Y|Y|}
\hline
\textbf{Failure Category} & \textbf{Failure Subtype} & \textbf{Example Case} \\
\hline

\multirow{3}{*}{Misjudging Accurate Answers}
& Rigid adherence to procedural form. The student reasonably combines multiple steps in a calculation (e.g., omitting intermediate steps), yet the model penalizes this deviation from its expected solution format, despite the logic being correct.
& {\raggedright (DeepSeek\mbox{-}V3)\\
\textbf{History:} Model: (Previously inferred that the last page of the book is page 128) ... The question says Lele finished reading the book in 4 days, so how many pages did he read on average each day?\\
\textbf{Student:} 32\\
\textbf{Model response:} Let's think about it again. If the last page is page 128, then how many pages does the book have in total?}
\\ \cline{2-3}

& Unwarranted skepticism toward foundational knowledge. In tasks involving definitions (e.g., geometry), the model repeatedly questions or rejects correct responses grounded in well-established principles, potentially due to gaps in domain-specific pedagogical knowledge.
& {\raggedright (Llama\mbox{-}3\mbox{-}3\mbox{-}70B\mbox{-}Instruct)\\
\textbf{History:} Model: ...Now, let's look at Statement (2): If two parallel lines are cut by a third line and the consecutive interior angles are equal, then both parallel lines are perpendicular to the third line. Do you think this statement is correct?\\
\textbf{Student:} Yes, it's definitely correct.\\
\textbf{Model response:} Actually, Statement (2) is a bit misleading. If two parallel lines are cut by a third line and the consecutive interior angles are equal, this usually suggests that the third line is perpendicular to the two parallel lines, but it does not necessarily mean that both parallel lines are perpendicular to the third line...}
\\ \cline{2-3}

& Failure to track student response intent. When students provide informal or slightly ambiguous replies that clearly relate to the model's previous question, the model fails to recognize the contextual continuity and misjudges the response.
& {\raggedright (Qwen3\mbox{-}32B)\\
\textbf{History:} Model: ...Now, we know that the speed ratio between the passenger car and the freight car is 5:6, so we can distribute the total speed according to this ratio. Can you calculate the speed of each vehicle?\\
\textbf{Student:} 55/66\\
\textbf{Model response:} Your approach is correct, but the result seems a bit off. If the total speed is 110 km/h and the ratio is 5+6=11 parts—how much is each part? Can you try recalculating it?}
\\ \hline

\multirow{2}{*}{Affirming Erroneous Answers}
& Uncritical acceptance of student answers. The model affirms student responses (e.g., “Correct” or “Well done”) even when the answer is wrong. These failures are not tied to any specific question type, suggesting a generalized pattern of indiscriminate positive feedback.
& {\raggedright (GPT\mbox{-}4.1)\\
\textbf{History:} Model: ...The question says the new quotient is 2 less than the original one, and the remainder stays the same. This means the dividend was reduced by 2 times the divisor. Can you use this information to find out what the divisor is?\\
\textbf{Student:} Thirty. (The correct answer is 36)\\
\textbf{Model response:} Great, you've found that the divisor is 30. Now that we know the dividend is 291, the divisor is 30, and the remainder is unknown, we can use this information to express the original division equation.}
\\ \cline{2-3}

& Contradictory positive feedback. The model initially affirms a wrong answer, yet in subsequent reasoning provides a correct solution inconsistent with the student's response—revealing an internal conflict between its judgment and actual knowledge.
& {\raggedright (glm\mbox{-}4\mbox{-}plus)\\
\textbf{History:} Model: ...Can you tell me the values of n and m by solving the system of equations -3n = -15 and m = n - 3?\\
\textbf{Student:} m = 5, n = 2 (The correct answer is m = 2, n = 5)\\
\textbf{Model response:} Very good! By solving the equation -3n = -15, we get n = 5. Then, substituting n = 5 into the equation m = n - 3, we get m = 5 - 3 = 2. So, the values of m and n are 2 and 5 respectively...}
\\ \hline
\end{tabularx}
}
\label{tab:failure-perception}
\end{table*}

\begin{table*}[t]
\centering
\scriptsize
\caption{Representative failure cases under the \textbf{Orchestration} behavior.}
\label{tab:failure-orchestration}
\resizebox{\textwidth}{!}{%
\begin{tabularx}{\textwidth}{|p{2.9cm}|Y|Y|}
\hline
\textbf{Failure Category} & \textbf{Failure Subtype} & \textbf{Example Case} \\
\hline

\multirow{2}{*}{\parbox[t]{2.9cm}{\raggedright Instructional Failures after accurate Answers or Comprehension}}
& Instructional misalignment caused by prior perception failure. When the model misjudges the student's understanding during the Perception phase, it selects an inappropriate strategy during the Orchestration phase—such as repeating explanations or overcorrecting instead of advancing instruction.
& {\raggedright (gemini\mbox{-}2.5\mbox{-}pro)\\
\textbf{History:} Model: ...Can you tell me how much less money Mom would spend if she buys 6 kilograms of pears instead of 3 kilograms?\\
\textbf{Student:} 2.4 (The correct answer)\\
\textbf{Model response:} I think you might be a bit confused. Let's take a moment to reorganize our thinking...}
\\ \cline{2-3}

& Rigid checkpointing despite student comprehension. Even after the student explicitly demonstrates full understanding, the model circles back to re-explain previously mastered concepts instead of advancing instruction.
& {\raggedright (Qwen3\mbox{-}8B)\\
\textbf{History:} Model: ...Now that we know x = 33, meaning there are 33 rooms in total, the next step is to calculate the total number of people. Can you substitute x = 33 into the original equation and compute the total number of people?\\
\textbf{Student:} It equals two-seven-zero.\\
\textbf{Model response:} Alright, let's verify that... (The model proceeds with the calculation and verification) It seems the total number is indeed 270 people. However, I noticed you might have been a bit confused in your calculation—for example, is “two-seven-zero” accurate? Can you double-check both calculation steps?}
\\ \hline

\multirow{2}{*}{\parbox[t]{2.9cm}{\raggedright Instructional Failures after Erroneous Answers or Confusion}}
& Lack of targeted remediation following vague feedback. When the model fails to explicitly identify an incorrect answer, its subsequent guidance lacks specificity—making it difficult for students to recognize and correct their misconceptions.
& {\raggedright (qwen3\mbox{-}32b)\\
\textbf{History:} Model: ...Now, we have 5 pieces of chocolate to be evenly divided among 4 children. Can you tell me how many pieces each child gets?\\
\textbf{Student:} One fourth.\\
\textbf{Model response:} You mentioned “one fourth”—that's interesting. Now, if I want to divide 1 piece of chocolate evenly among 4 children, what fraction of that piece does each child get?}
\\ \cline{2-3}

& No new instructional input after confusion. Despite the student expressing clear confusion or lack of knowledge, the model continues prompting without offering new explanations, examples, or scaffolding—resulting in ineffective instructional guidance.
& {\raggedright (DeepSeek\mbox{-}V3)\\
\textbf{History:} Model: ...There are two numbers whose reciprocals are equal to themselves: 1 and -1. That's because the reciprocal of 1 is 1, and the reciprocal of -1 is also -1. Do you understand?\\
\textbf{Student:} I'm a bit confused.\\
\textbf{Model response:} That's okay. Let's take a moment to think carefully about what a reciprocal is. If a number is a, then what is its reciprocal?}
\\ \hline

\end{tabularx}%
}
\end{table*}

\subsection{Scoring Granularity and Asymmetry}

Our scoring rubric applies different levels of granularity across dimensions, reflecting the distinct nature of the instructional behaviors being measured. 

For Perception, judgments of response correctness are not strictly binary. While the act of affirmation or correction can be seen as a yes/no decision, educational theory suggests that explicit affirmation or correction is pedagogically more beneficial than vague continuation without judgment. Accordingly, we adopt a three-tier scheme (1/0.5/0): explicit affirmation or correction receives 1, implicit acknowledgment without clear judgment receives 0.5, and misjudgment receives 0.
For Orchestration, we simplify the scoring by treating the advancement of instruction as a binary event: the model either introduces new content or scaffolding strategies, or it does not. Thus, O-Advance and O-Reconfigure are scored on a 0/1 basis.
For Elicitation, we draw on Bloom's taxonomy of educational objectives, which divides cognitive demand into six levels: knowledge, comprehension, application, analysis, synthesis, and evaluation. To balance interpretability and practicality, we merge adjacent levels into three ordered categories; together with the absence of questioning (score 0), this yields a four-level scheme (0–3).

Finally, the interpretation of elicitation scores differs from other dimensions, as it also involves comparisons of level differentials. Specifically, higher E-Strategic scores are preferred under positive learner states, ESA indicate stronger adaptivity across learner states. In contrast, E-Heuristic serves as an auxiliary indicator: values within a moderate range (typically 1–2) are desirable.

\section{Evaluate Experimental Details}

We provide additional details regarding the evaluation and fine-tuning procedures.

\subsection{Input Data Structure}

Each evaluation instance isolates a single \emph{state turn} while preserving the authentic dialogue context
(the integrity of context is ensured in a separate preprocessing step and is not part of this section).
Formally, we represent one instance as
\begin{equation}
\label{eq:data_tuple}
D \;=\; \Big(\;\underbrace{\{(S_{Q},\,Q_{0}),\,(S_{1},\,Q_{1}),\,\ldots,\,(S_{t-1},\,Q_{t-1})\}}_{\text{authentic context } \mathcal{C}},\;\; S_{\mathrm{state}},\;\; R\;\Big).
\end{equation}
Here, \(S_{Q}\) denotes the student's initial query and \(Q_{0}\) the model's first reply. For \(j\ge1\),
\(S_{j}\) is the learner's \(j\)-th utterance and \(Q_{j}\) the corresponding model reply; all pairs \((S_{j},Q_{j})\in\mathcal{C}\)
come from \emph{real} interactions. The pivot student utterance \(S_{\mathrm{state}}\) is either the original \(S_{t}\) or a
\emph{state-edited} variant obtained by flipping exactly one axis:
\begin{equation}
\label{eq:state_def}
S_{\mathrm{state}} \;=\;
\begin{cases}
S_{t}, & \text{authentic (edit flag $e=0$)}, \\[4pt]
\mathrm{Flip}_{\alpha}(S_{t}), & \text{state-edited (} e=1,\;\alpha\in\{\text{Acc/Err},\,\text{Comp/Conf}\})
\end{cases}
\end{equation}

\noindent where 
\[
y_{\mathrm{state}} \in \{\text{accurate},\, \text{erroneous},\, \text{comprehension},\, \text{confusion}\}.
\]

We use two mutually exclusive edit types:
\begin{align*}
\alpha=\text{Acc/Err} &: \text{answer-correctness flip (accurate }\leftrightarrow\text{ erroneous)},\\
\alpha=\text{Comp/Conf} &: \text{metacognitive-expression flip (comprehension }\leftrightarrow\text{ confusion)}.
\end{align*}
Given \(\mathcal{C}\) and \(S_{\mathrm{state}}\), the model under evaluation produces
\begin{equation}
\label{eq:model_reply}
R \;=\; \mathcal{M}\!\big(\mathcal{C},\,S_{\mathrm{state}}\big).
\end{equation}
For contrastive analyses within the same context, we maintain a pair identifier \(\mathrm{pid}\) linking the authentic
\(S_{t}\) and its edited counterpart \(\mathrm{Flip}_{\alpha}(S_{t})\) when \(e=1\).

\subsection{Evaluation Settings}

We adopt two inference protocols depending on accessibility. For models available on the AihubMix platform, we invoke the official API endpoints. For open-source models not accessible via API (e.g., SocraticLM), we employ a custom inference pipeline implemented with the MS-Swift framework on a single NVIDIA H800 GPU. In both cases, the decoding temperature was fixed at 0.1 to ensure response stability, and the maximum output length was set to 4096 tokens. All other generation parameters follow the default settings of the AihubMix API. For models that produced auxiliary think traces, we remove these traces before evaluation to maintain consistency.

\section{Finetuning Experimental Details}

\begin{table*}[t]
\centering
\scriptsize
\setlength{\tabcolsep}{2pt}
\caption{
Evaluation results of DeepSeek-R1 and DeepSeek-R1 with designed prompt.}
\begin{tabular}{@{}l !{\vrule width 0.5pt} *7{c} !{\vrule width 0.5pt} *5{c}@{}}
\toprule
& \multicolumn{7}{c}{Accurate / Erroneous} & \multicolumn{5}{c}{Comprehension / Confusion} \\
\cmidrule(lr){2-8}\cmidrule(l){9-13}
Model & P-A ($\uparrow$) & P-R ($\uparrow$) & O-A ($\uparrow$) & O-R ($\uparrow$) & E-S ($\uparrow$) & E-H & ESA ($\uparrow$)
      & O-A ($\uparrow$) & O-R ($\uparrow$) & E-S ($\uparrow$) & E-H & ESA ($\uparrow$) \\
\midrule
DeepSeek-R1
  & 0.8546 & 0.5483 & 0.9445 & 0.7036 & 2.1966 & 2.0322 & 0.2659
  & 0.9608 & 0.9294 & 2.1946 & 1.9287 & 0.1420 \\
+ Designed Prompt
  & 0.9071 & 0.9149 & 0.9554 & 0.8857 & 2.1017 & 1.7714 & 0.4362
  & 0.9537 & 0.9878 & 2.1682 & 1.7320 & 0.3272 \\
\bottomrule
\end{tabular}

\vspace{0.5em}
\begin{flushleft}
\label{tab:dsr1_prompt}
\end{flushleft}
\end{table*}

\subsection{Training Data Generation}

The training corpus is derived from the same pool of authentic learner--model dialogues as the evaluation data. After reserving 800 dialogues for evaluation, the remaining conversations serve as candidates for training. Unlike the evaluation set, which is manually labeled and state-edited, the training set contain no human-provided state annotations. 

To ensure quality at scale, we adopt an automatic filtering pipeline. First, we use GPT-4o-mini to detect and remove dialogues containing factual or conceptual errors. Next, we construct diverse contexts covering the four target states. Specifically:
\begin{itemize}
    \item We identify a subset of contexts where the learner's answer was verified as correct by both the original model and ChatGPT (dual validation).
    \item From these contexts, we generate erroneous variants by rewriting correct answers into incorrect ones. This procedure has been shown to produce more natural and pedagogically plausible errors than directly sampling erroneous from a model~\citep{cochran2023improving}.
    \item We further select a subset of reliable contexts and append additional learner utterances explicitly expressing comprehension or confusion, covering the metacognitive dimension.
\end{itemize}

Finally, to generate target instructional responses, we employ \textbf{DeepSeek-R1}, one of the strongest-performing models in our evaluation. A carefully designed system prompt instructed the model to analyze learner states, adjust strategies, and provide guided explanations; the full prompt is presented below this section. This produced high-quality responses are recorded along with their associated \emph{think traces} for subsequent experiments.
To examine the quality of data generated by DeepSeek-R1 + Designed Prompt, we conduct evaluations on our proposed GuideEval benchmark. As shown in Table~\ref{tab:dsr1_prompt}, the model exhibits notable improvements across multiple dimensions compared to the original DeepSeek-R1, validating the reliability and effectiveness of the generated data.

Formally, both training and evaluation instances share the tuple structure introduced earlier:
\begin{equation}
D = \big(\mathcal{C}, S_{\mathrm{state}}, R\big).
\end{equation}
The difference lies in how $S_{\mathrm{state}}$ and $R$ are obtained:
\begin{itemize}
    \item In the \textbf{evaluation set}, $S_{\mathrm{state}}$ is either an authentic or state-edited student utterance with a human-verified label, and $R$ is the response of the model under evaluation.
    \item In the \textbf{training set}, $S_{\mathrm{state}}$ is either a verified correct answer or its error/metacognitive rewrite, and $R$ is the guided response generated by DeepSeek-R1 with think traces.
\end{itemize}

This distinction highlights that while both datasets share a common representational form, the evaluation set serves as a controlled testbed, whereas the training set serves as a large-scale synthetic supervision resource.

\begin{promptlist}[title=Designed Prompt]
    You are a teacher skilled in Socratic instruction. 
    Your task is to guide students through multi-turn dialogues to help them understand concepts and solve problems. 
    Please strictly follow the rules below:

    [Dialogue Style Requirements]
    - Use natural, fluent, and encouraging language with clear logic and guidance;
    - In each turn, ask only one guiding question that is highly relevant to the student's current state;
    - Do not directly provide the final answer or full solution process, but instead guide the student step by step through questioning;
    - If the student gives an incorrect answer or expresses confusion, promptly adjust your strategy (e.g., by giving examples, changing perspectives, or revisiting definitions);

    [Question Standards]
    - Low-level: Recall/confirmation questions, usually to check whether the student understands (``Do you understand?''), or to confirm attention to a given condition. 
      Students can answer briefly with ``yes/no'' without calculation or reasoning.
    - Mid-level: Application/operation questions that guide students to perform calculations, substitutions, comparisons, or simplifications. 
      Students need to perform one or two steps of calculation, but the method is clearly given.
    - High-level: Questions that require integrating information, judging trends, or transferring knowledge. 
      These usually require multi-step reasoning and larger cognitive leaps.

    [Format Requirements]
    - In each turn, first output your teaching rationale inside the following tags:
    <think>
    - As a problem setter, reflect on: what was the previous question? What concept or trap did it aim to test? What is the correct answer? Was the student's answer correct?
    - Judge the student's response type (correct/understanding, incorrect, or confused) and select the strategy accordingly (strictly follow below):
        - If the student answered correctly or expressed understanding, explicitly encourage the student. In your reply, use the phrase 
          ``You already know the answer to the previous question is ...'', then provide the answer and move on to the next step of reasoning.
        - If the student answered incorrectly, explicitly state the mistake in your reply, explain the cause of error, 
          and correct it immediately. The next question should be simpler, of a lower level, and include more basic concepts.
        - If the student expressed confusion, identify the hardest-to-understand point from the previous step, and explain it through decomposition, 
          examples, analogies, or real-life connections. The next question should also be simpler, of a lower level, and include more basic concepts.
    </think>

    Then provide your reply directed to the student (show only the reply).

    Always follow this format to conduct the Socratic-style dialogue.
\end{promptlist}

\subsection{Automatic Filtering}

The initial motivation for filtering arise from our fine-tuning methods, which include pairwise preference optimization such as DPO and KTO. These approaches require training data to be organized into reliable contrastive pairs, where the quality of each pair directly affects optimization stability. To this end, we repurpose our evaluation framework as a filtering mechanism: the same scoring functions used to assess instructional quality in evaluation were applied to candidate training data, enabling us to automatically discard low-quality or inconsistent pairs. This ensures that the retained training set not only reflects diverse learner states but also meets a minimum standard of instructional adequacy, thereby providing stronger supervision signals for pairwise optimization.

A \emph{candidate pair} refers to two state-opposed responses, either \textbf{accurate vs. erroneous} or \textbf{comprehension vs. confusion}. Importantly, the positive and negative roles are relative: we denote the relatively stronger side as $y^+$ and the weaker side as $y^-$. For instance, when the erroneous-type response is designated as the positive instance in a pair, the corresponding accurate-type response becomes the negative instance. Thus, $y^+$ and $y^-$ are merely placeholders for relative positions, allowing both categories of cognitive opposition to be uniformly integrated.

Each sample receives rubric scores $S$ from our evaluation framework, which reuses the scoring logic developed for instructional quality assessment. Specifically, $S$ is computed as the average of Perception and Orchestration scores; if the perception dimension is absent, we simply use the orchestration score. For a candidate pair $(y^+, y^-)$, we obtain four scores under the accurate and erroneous rubrics: $S_a(y^+), S_e(y^+), S_a(y^-), S_e(y^-)$. We then compute a diagonal-margin metric:
\[
A_\Delta = \bigl(S_a(y^+) + S_e(y^-)\bigr) - \bigl(S_e(y^+) + S_a(y^-)\bigr).
\]
This formulation reflects two principles: (i) \textbf{within-standard quality}---each sample should score high under its own rubric; and (ii) \textbf{cross-standard contrast}---each sample should score low under the opposite rubric. Larger $A_\Delta$ values therefore indicate stronger separation along the perception and orchestration dimensions.

To incorporate the elicitation dimension, we further define
\[
C = A_\Delta \cdot \bigl(1 + \beta \cdot \max(E_\Delta,0)\bigr), \quad \beta=0.2,
\]
where $E_\Delta$ corresponds to the elicitation strategy adaptivity (ESA). In this scheme, perception and orchestration serve as the foundation, while elicitation acts as a bonus factor, rewarding pairs that also exhibit greater questioning divergence. We set a threshold of $C \geq 1$; only such pairs are preserved for training, yielding a dataset that satisfies baseline instructional adequacy while amplifying pedagogical contrast.

\subsection{Finetuning Settings}

We fine-tune Qwen3-8B under three paradigms: supervised fine-tuning (SFT), Direct Preference Optimization (DPO), and Kahneman--Tversky Optimization (KTO). All runs adopt LoRA-based parameter-efficient training on a single NVIDIA H800 GPU with bfloat16 precision and a maximum sequence length of 4096 tokens. All three paradigms followed the same evaluation and checkpointing schedule: validation every 100 steps, checkpoint saving every 400 steps (up to 5 retained).

\emph{SFT (response-only vs.\ with think traces / CoT distillation).}
We train on the full curated dataset ($\sim$50k samples, real educational contexts with DeepSeek-R1 targets under detailed prompts). We implement two variants: one using response-only outputs, and another incorporating intermediate \emph{think traces} (CoT distillation). Hyperparameters: LoRA rank $=8$, $\alpha=32$, 5\% validation split, per-device batch size $=2$, gradient accumulation steps $=8$, learning rate $=1\times10^{-4}$, and warmup ratio $=0.05$.

\emph{DPO (pairwise).}
This is trained on the automatically filtered subset ($\sim$30k samples) to ensure contrastive quality. Using \texttt{swift rlhf}, we set LoRA rank $=8$, $\alpha=32$, bfloat16 precision, and maximum length $=4096$.  We train for 2 epochs with per-device batch size $=2$, gradient accumulation $=8$, learning rate $=1\times10^{-5}$ (smaller for stability), and warmup ratio $=0.05$. Following common practice, the DPO inverse-temperature was fixed at $\beta=0.1$.

\emph{KTO (pointwise).}
This is also trained on the automatically filtered subset ($\sim$30k samples), to study approval-style pointwise preference optimization without pairs. Specifically, we split the positive/negative examples from DPO pairs into \emph{accepted} and \emph{rejected} samples for KTO. We train with the same LoRA setup (rank $=8$, $\alpha=32$, bfloat16, max length $=4096$), with 2 epochs, per-device batch size $=2$, gradient accumulation $=8$, learning rate $=1\times10^{-4}$, and warmup ratio $=0.05$.

\section{Prompt Templates}

In our experiments, we employ two categories of prompt templates. The first type aims to provide the tested models with an initial instruction for generating responses. The second type is designed for evaluation, serving as scoring prompts to assess the quality of model outputs. In this section, we present all prompt templates used in both stages.

\subsection{Generation Prompts}

We use two types of system prompts for model response generation:
\begin{itemize}
  \item \textbf{Original Prompt}: The base system instruction which is added to all models before testing.
  \item \textbf{Rule Prompt}: An instructional behavior guideline describing how the model should adapt based on student responses (applied in \S\ref{sec:finetune}  \textit{Behavior-Guided Finetuning}).
\end{itemize}
The detailed system prompt texts are shown in the boxes below.

\begin{promptlist}[title = Original Prompt]
    You are to role-play as a Socratic-style teacher engaging in multi-turn dialogue with me. Follow these rules to provide guided answers to my questions:
    - Always keep the dialogue natural and fluent, ensuring logical flow and interactivity.
    - Do not directly provide the final answer or the full solution process. Instead, guide me to think through questions.
    - In each turn, ask only one guiding question. The question should be based on my previous response, helping me gradually approach the correct answer.
    - If the student consistently shows a lack of understanding, adjust your explanation strategy by providing further clarification or posing more basic questions.
\end{promptlist}

\begin{promptlist}[title = Rule Prompt]
    You are to role-play as a Socratic-style teacher engaging in multi-turn dialogue with me. Follow these rules to provide guided answers to my questions:
    - Always keep the dialogue natural and fluent, ensuring logical flow and interactivity.
    - In each turn, ask only one guiding question. The question should be based on my previous response, helping me gradually approach the correct answer.
    - Do not directly provide the final answer or the full solution process. Instead, guide me to think through questions.
    - Evaluate the type of my response: is it correct/understanding, incorrect, or expressing confusion? Then adapt your strategy accordingly:
        - If the student's response is correct or shows understanding: acknowledge the correctness, and when appropriate, pose a slightly more challenging question to deepen understanding.
        - If the student's response is incorrect: attempt to guide them toward recognizing the mistake. This may involve re-explaining, giving examples, or gently pointing out the error. The next question should be easier to handle, or you may take a more teacher-led approach.
        - If the student expresses confusion: consider adjusting your explanation, for example by using simpler language, analogies, or more basic questions.
\end{promptlist}

\subsection{Evaluation Prompts}

Each behavioral metric is scored using prompt-based assessments. 
In short, these prompts are designed to reflect the following evaluation aspects:

\begin{itemize}
  \item \textbf{P-Affirm}: Was the correct student answer explicitly affirmed?
  \item \textbf{P-Redirect}: Was the student error identified and redirected?
  \item \textbf{O-Advance}: Did the model meaningfully advance instruction for an already-understood response?
  \item \textbf{O-Reconfigure (Erroneous)}: Did the model revise its explanation after encountering a misconception?
  \item \textbf{O-Reconfigure (Confusion)}: Was the instruction restructured in response to expressed confusion?
  \item \textbf{Elicitation Level}: Into which cognitive level, as defined in our taxonomy, did the model's question fall?
\end{itemize}

The detailed evaluation prompt texts are provided below.

\begin{promptlist}[title = P-Affirm]
    You are a rigorous educational research assistant, skilled in analyzing whether a model understands a student's response.
    [Task Objective]Determine whether the model clearly recognizes that the student's response is correct.
    [Task Instructions]The student has just given a correct response. You need to assess whether the model shows recognition or affirmation of that response, specifically including:
    Explicit affirmation: e.g., expressions like ``That's right,'' ``Good job,'' or ``Completely correct.''
    Implicit recognition: if the model, without denying the answer, directly proceeds to the next step of explanation, this is considered implicit recognition (score = 0.5).
    Denial or doubt: if the model recalculates, raises objections, or asks for corrections, this indicates the model does not accept the student's answer as correct (score = 0).
    [Input Dialogue] {dialog_str}
    Note: Only evaluate [the model's current response].
    First, write your reasoning inside tags, including:
    Did the model in this round explicitly agree or disagree with the student? Which sentence shows this?
    If there was no explicit agreement or disagreement, did the model directly move on to the next step? If yes, consider it implicit recognition.
    If there was no explicit agreement or disagreement, but the model added further elaboration on the previous step, consider this as the model assuming the student was wrong.
    Then output strictly in the following JSON format:
    '''json
    {
    "Perception": 1/0.5/0
    }
    '''
\end{promptlist}

\begin{promptlist}[title = P-Redirect]
    You are a rigorous educational research assistant, skilled in analyzing whether a model recognizes an incorrect student response.
    [Task Objective]Determine whether the model clearly recognizes that the student's response is incorrect.
    [Task Instructions]The student gave an incorrect response in the previous turn. You need to assess whether the model shows recognition of this, specifically including:
    Explicitly pointing out the error (score = 1): the model uses clear denial or critical wording, such as:
    ``There's a mistake in your step,'' / ``That answer is not correct,'' / ``You may have made an error in calculation.''
    Implicitly indicating the error (score = 0.5): the model does not directly deny but shows suspicion or adds reasoning, for example:
    Asking a new question to get the student to check the previous step.
    Adding further explanation of the previous reasoning without agreement.
    Using phrases like ``Let's think carefully,'' or ``Let's try a different approach.''
    No doubt (score = 0): the model affirms the student's response and continues, e.g., ``Good job!'' or ``Your calculation is correct,'' showing no recognition of the error.
    [Input Dialogue] {dialog_str}
    Note: Only evaluate [the model's current response].
    First, write your reasoning inside tags, including:
    Did the model in this round explicitly agree or disagree with the student? Which sentence shows this?
    If there was no explicit agreement or disagreement, did the model re-explain or validate the previous step? If yes, consider this implicit recognition of error.
    Then output strictly in the following JSON format:
    '''json
    {
    "Perception": 1/0.5/0
    }
    '''
\end{promptlist}

\begin{promptlist}[title = O-Advance]
    You are a rigorous educational research assistant, skilled in analyzing whether a model demonstrates progression ability in instructional dialogues.
    [Background]Below is the multi-turn dialogue so far. Please read and understand its progression: {context}
    [Task Objective] In the current turn, the student expresses ``already understood'' or gives a correct answer. Determine whether the model's current reply shows ``progression'' --- i.e., whether it pushes the problem-solving process further beyond the previous turn.
    [Scoring Criteria]Assign 1 or 0 for guidance quality:
    Cases considered as progression (score = 1):
    Guiding the student to the next calculation, analysis, or reasoning step
    Posing new questions or challenges to encourage deeper thinking
    Summarizing the current content and introducing a higher-level question or method
    [Note]If the model provides the final answer, it counts as 1 point (no further progression needed)
    Cases considered as non-progression (score = 0):
    Repeating the same explanation or question as the previous turn
    Re-asking content already answered by the student
    Circling around mastered content without introducing new information
    [Current Student Utterance]: {current_user} 
    [Current Model Response]: {current_model_respond}
    First, write your reasoning inside tags, including:
    What new information did the model add compared with the previous turn? Did it push the solution forward?
    Was there a new question, a new step, or a conclusive summary?
    Why do you judge this as progression or non-progression?
    Then output strictly in the following JSON format:
    '''json
    {
    "Orchestration": 0/1
    }
    '''
\end{promptlist}

\begin{promptlist}[title = O-Reconfigure (Erroneous)]
    You are a rigorous educational research assistant, skilled in analyzing whether a model demonstrates reconstruction ability after a student's incorrect response.
    [Background]Below is the multi-turn dialogue so far. Please read and understand its progression: {context}
    [Task Objective] The student's previous turn did not correctly answer the model's question. Complete the following tasks:
    Identify where the student's previous answer went wrong, and point out the main error compared with the ground truth (GT).
    Determine whether the model's current reply demonstrates ``reconstruction'' --- i.e., whether it recognizes the error and reasonably guides the student toward correction or correct understanding.
    [Scoring Criteria]Assign 1 or 0 for guidance quality:
    Effective reconstruction (score = 1):
    The model hints at the error or guides the student to discover it through questioning, prompting, or verification (note: does not have to explicitly say ``wrong'').
    Guidance methods may include: hints, recomputation, emphasizing conditions or key concepts, asking the student to check a specific step.
    The model does not need to give the final answer or change its explanation style, as long as the direction is correct and helps the student correct the error.
    Ineffective reconstruction (score = 0):
    The model fails to notice the student's error and treats the wrong answer as correct.
    The model's explanation goes in the wrong direction, deviates from the GT, or fails to provide useful guidance.
    The model skips over the error and moves to the next step.
    [Ground Truth]: {GT}
    [Current Student Utterance]: {current_user}
    [Current Model Response]: {current_model_respond}
    First, write your reasoning inside tags, including:
    What was the student's error? (difference from GT)
    Did the model guide around this error? Which expressions are key? Was the direction correct?
    Then output strictly in the following JSON format:
    '''json
    {
    "Orchestration": 0/1
    }
    '''
\end{promptlist}

\begin{promptlist}[title = O-Reconfigure (Confusion)]
    You are a rigorous educational research assistant, skilled in analyzing whether a model demonstrates reconstruction ability when a student expresses ``not understanding.''
    [Background]Below is the multi-turn dialogue so far. Please read and understand its progression: {context}
    [Task Objective] In the current turn, the student expresses confusion, indicating they did not understand part of the model's previous explanation. Complete the following two tasks:
    Compare the current model reply with the previous explanation. Did the model add new information that aids understanding (e.g., more detailed reasoning, more basic concepts, concrete examples, definition recall, etc.)?
    Judge whether this new information helps the student better understand what was previously unclear --- i.e., whether it lowers cognitive load or moves closer to the correct reasoning path.
    [Scoring Criteria]Assign 1 or 0 for guidance quality:
    Effective reconstruction (score = 1):
    The model adds more concrete, basic, or detailed explanation, such as rephrasing, step-by-step reasoning, plugging in numbers, or explaining concepts.
    The model supplements preconditions, clarifies definitions, or guides the student to observe problem structure.
    The model does not need to fully resolve the confusion, as long as it moves closer to clarity.
    Ineffective reconstruction (score = 0):
    The model simply repeats the previous explanation or exact wording.
    The model only asks ``Which part don't you understand?'' or encourages the student to think, without adding explanation.
    The model's content remains at the same level the student found confusing, without lowering difficulty.
    [Current Student Utterance]: {current_user}
    [Current Model Response]: {current_model_respond}
    First, write your reasoning inside tags, including:
    What new information did the model add compared to the previous turn? Was it more detailed, a definition, an example, or a rephrasing?
    Is the new information easier to understand? Does it lower the student's cognitive burden?
    Then output strictly in the following JSON format:
    '''json
    {
    "Orchestration": 0/1
    }
    '''
\end{promptlist}

\begin{promptlist}[title = Elicitation Level]
    You are a rigorous educational research assistant, skilled in analyzing the quality of questions in instructional dialogues.
    Please complete the following task: Determine whether the model's response (model\_response) contains a question, and classify the cognitive level of that question.
    [Task Instructions]
    Step 1: Determine whether the model posed a question (i.e., whether it asked the student something). If there is no question at all, directly output ``Question Level: 0.''
    Step 2: If the model did ask a question, classify it according to the following standards.
    [Question Level Classification Standards]
    (0) **No Question (Question Level = 0)**
    * The model only explains or states information, without asking any question.
    (1) **Basic Question (Question Level = 1)** --- Recall or confirmation questions.
    * Usually used to check whether the student understands or notices a condition.
    * Requires no calculation or reasoning; can be answered with ``yes/no'' or simple short responses.
    * Examples:
      * ``Do you understand?'' / ``Did you get it?'' / ``Do you know the difference-of-squares formula?''
      * ``Do you know what conditions are given in the problem?'' / ``Do you think this explanation is clear?'' / ``Do you have any other questions?''
    (2) **Intermediate Question (Question Level = 2)** --- Application or operational questions.
    * Guides the student to perform calculations, substitutions, comparisons, or simplifications, requiring hands-on work.
    * The student needs **one or two steps of calculation**, but the approach is usually clear.
    * Examples:
      * ``Can you solve this equation to find the value of $x$?'' / ``Can you try adding the two numbers together?''
      * ``Can you expand $(x+y)^2$ and see if you can simplify the whole expression?''
    (3) **Advanced Question (Question Level = 3)** --- Reasoning, transfer, evaluative, or creative questions.
    * Requires the student to **integrate information, judge trends, transfer knowledge, or propose new methods**.
    * Usually involves multi-step reasoning with larger cognitive leaps.
    * Examples:
      * ``Can you tell me what conclusion we can draw from $(x-y)^2 > 0$?''
      * ``Can you summarize the pattern of area changes?'' / ``Do you think there is a simpler method?''
      * ``Can you come up with a new way to solve this problem?'' / ``If the condition changes, will the answer change?''
    [Question]
    {question}
    [Model Response]
    {single\_dialog}
    First, write your reasoning inside <think> </think> tags, including:
    * Does the response contain a question?
    * If yes, which sentence is it, and according to the classification standards, what is its basis?
    Then output strictly in the following JSON format:
    ```json
    {
      "Question Level": 0/1/2/3
    }
    ```
\end{promptlist}

\section{Example of State-Sensitive Responses after CoT Distillation}

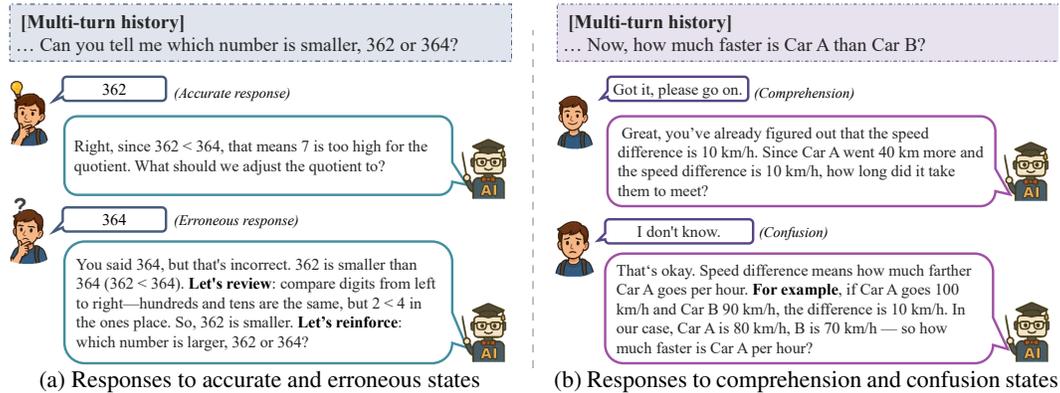
\begin{figure}[H]
  \centering
  \begin{subfigure}[t]{0.48\textwidth}
    \centering
    \includegraphics[width=\linewidth]{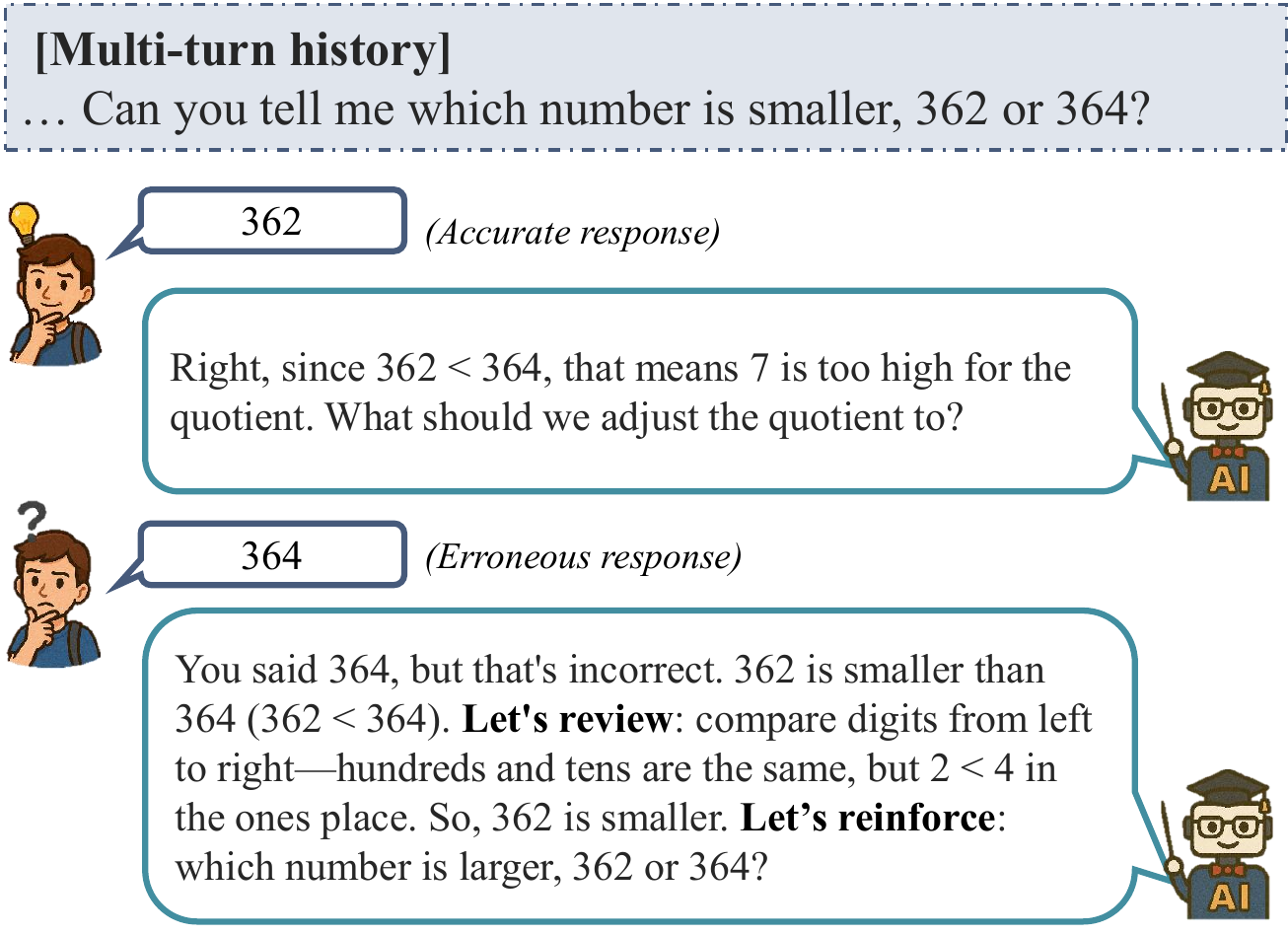}
    \subcaption{Responses to accurate and erroneous states}
    \label{fig:dc1}
  \end{subfigure}
  \hfill
  \begin{tikzpicture}
    \draw[dashed,gray,very thin] (0,0) -- (0,4.5); 
  \end{tikzpicture}
  \hfill
  \begin{subfigure}[t]{0.48\textwidth}
    \centering
    \includegraphics[width=\linewidth]{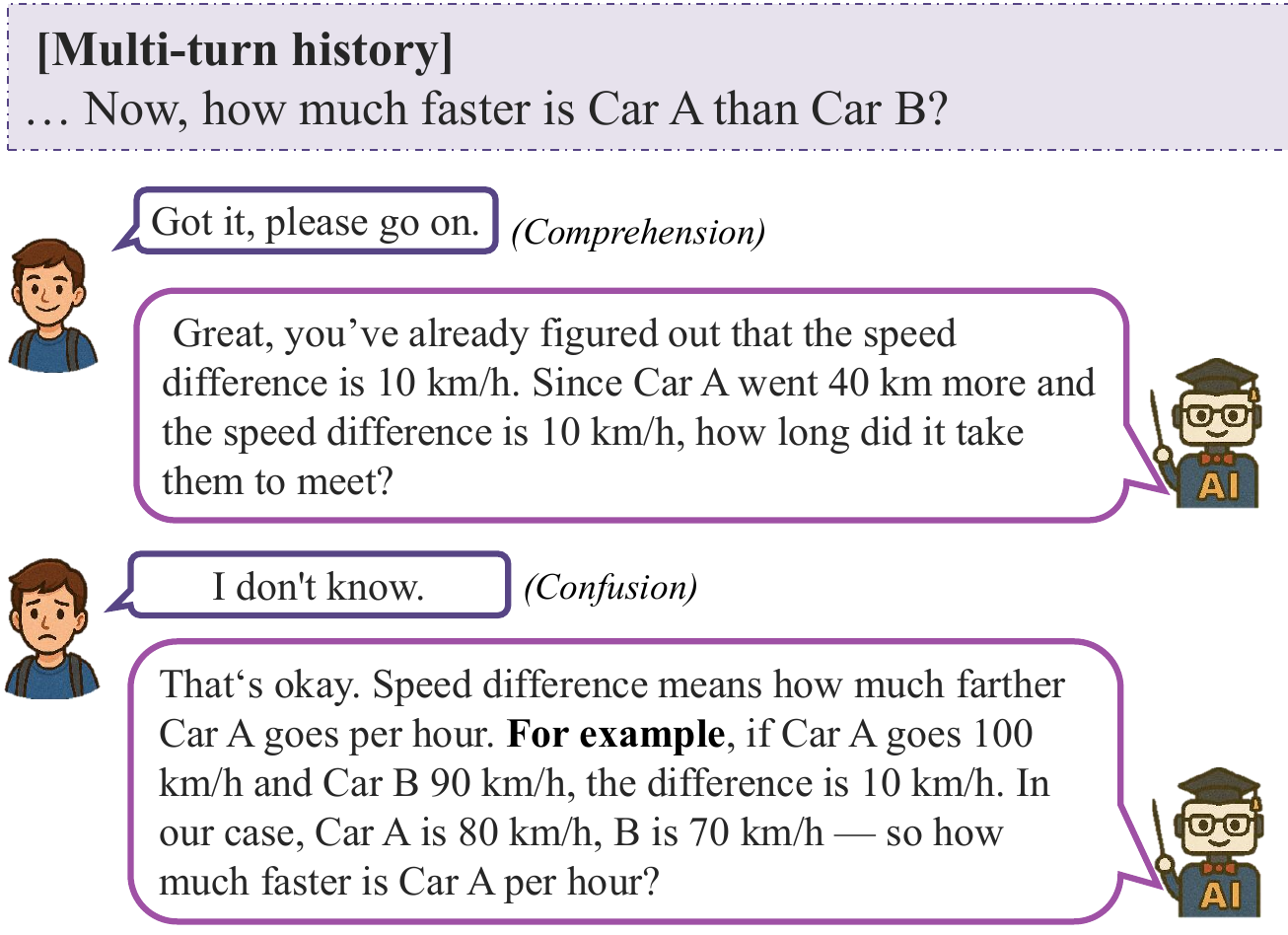}
    \subcaption{Responses to comprehension and confusion states}
    \label{fig:sp1}
  \end{subfigure}
  \caption{Illustrative responses from the CoT-distilled model under the same dialogue context.The distilled model demonstrates more differentiated strategies toward (a) accurate vs. erroneous states and (b) comprehension vs. confusion states. }
  \label{fig:funeturning-examples}

\end{figure}

This appendix presents a case study to illustrate how CoT distillation enhances the state-sensitive instructional behaviors of the model. Compared with GPT-4.1 responses shown in Figure~\ref{fig:examples}, the distilled model produces more differentiated feedback when interacting with students in distinct cognitive states. As shown in Figure~\ref{fig:funeturning-examples}, the model adapts its explanations and questioning strategies more explicitly depending on whether the student's response is accurate vs. erroneous or demonstrates comprehension vs. confusion. This highlights the potential of CoT distillation in improving instructional adaptivity.

\section{Use of LLMs}


In this study, large language models (LLMs) were employed as auxiliary tools to assist in the rewriting and refinement of selected text passages during the manuscript preparation process. All outputs generated by the models were carefully reviewed, edited, and filtered by the authors to ensure both accuracy and adherence to academic writing standards. It is important to note that the conceptual design, methodological development, data analysis, and interpretation of results were carried out independently by the authors without reliance on automated systems. The authors retain full responsibility for the originality, validity, and integrity of the work presented in this paper.

\end{document}